\documentclass{article}
\pdfoutput=1

\PassOptionsToPackage{numbers, compress}{natbib}

\usepackage[preprint]{neurips_2020}

\usepackage{times}
\usepackage{graphicx}
\usepackage[caption=false]{subfig}
\usepackage{amsmath,amssymb,amsthm}
\usepackage{epstopdf}
\usepackage{bm,xspace}
\usepackage{verbatim}
\usepackage{multirow}
\usepackage{wrapfig}
\usepackage{balance}
\usepackage{booktabs}
\usepackage{calc}

\newcommand{\ra}[1]{\renewcommand{\arraystretch}{#1}}

\def\bb{\mathbf{b}}

\def\ee{\mathbf{e}}

\def\kk{\mathbf{k}}

\def\qq{\mathbf{q}}

\def\vv{\mathbf{v}}

\def\yy{\mathbf{y}}
\def\zz{\mathbf{z}}

\def\AA{\mathbf{A}}

\def\GG{\mathbf{G}}

\def\KK{\mathbf{K}}
\def\LL{\mathbf{L}}

\def\QQ{\mathbf{Q}}

\def\VV{\mathbf{V}}

\def\YY{\mathbf{Y}}
\def\ZZ{\mathbf{Z}}

\def\D{\mathcal{D}}
\def\E{\mathcal{E}}
\def\F{\mathcal{F}}
\def\G{\mathcal{G}}

\def\N{\mathcal{N}}

\def\P{\mathcal{P}}

\def\V{\mathcal{V}}

\def\Re{\mathbb{R}}

\def\btheta{{\bm\theta}}

\newcommand{\paren}[1]{\left(#1\right)}
\newcommand{\abs}[1]{\left\lvert#1\right\rvert}

\newcommand{\ceil}[1]{\left\lceil#1\right\rceil}

\usepackage[utf8]{inputenc}
\usepackage[T1]{fontenc}
\usepackage{hyperref}
\usepackage{url}
\usepackage{amsfonts}
\usepackage{nicefrac}
\usepackage{microtype}
\usepackage{wrapfig}

\title{Auto-decoding Graphs}

\author{
  Sohil Atul Shah \\
  Intel Labs \\
  \And
Vladlen Koltun \\
Intel Labs \\
}

\begin{document}

\maketitle

\begin{abstract}
We present an approach to synthesizing new graph structures from empirically specified distributions. The generative model is an auto-decoder that learns to synthesize graphs from latent codes. The graph synthesis model is learned jointly with an empirical distribution over the latent codes. Graphs are synthesized using self-attention modules that are trained to identify likely connectivity patterns. Graph-based normalizing flows are used to sample latent codes from the distribution learned by the auto-decoder. The resulting model combines accuracy and scalability. On benchmark datasets of large graphs, the presented model outperforms the state of the art by a factor of 1.5 in mean accuracy and average rank across at least three different graph statistics, with a 2x speedup during inference.
\end{abstract}

\section{Introduction}
\label{sec:Introduction}

Generative models of graphs are widely used to model social, biological, and digital networks~\citep{Newman2018}. These models are used throughout science and engineering, for example to generate realistic graphs for training and benchmarking algorithms and for evaluating hypotheses. Classic models are based on statistical and structural assumptions, such as power-law distributions, small worlds, and others~\citep{erdds1959random,WattsStrogatz1998,Barabasi1999,leskovec2010kronecker}. However, a growing body of empirical evidence suggests that these assumptions do not capture the complexity of real-world networks~\cite{JacksonRogers2007,Lima-Mendez2009,BroidoClauset2019}. Graph synthesis models based on these assumptions may therefore not faithfully represent real data.

Recent advances in generative modeling and graph neural networks have enabled the development of \emph{empirical} graph synthesis models. These are richly parameterized models that are trained directly on data (exemplars of graphs). The models learn to synthesize graphs that match the empirical structure of observed data.
Early efforts in this direction made structural assumptions that limited the models to synthesis of small graphs (up to 50--100 nodes), constrained them to learning from a single exemplar, or limited their accuracy~\citep{kipf2016variational,li2018learning,liu2018constrained,Simonovsky2018}.

More recently, sequential generation models have been developed that can scale to large graphs with thousands of nodes~\citep{you2018graphrnn,liao2019gran}. These models alleviate the scalability issues of earlier work, but come with their own limitations. The best-performing variant of the GraphRNN model~\cite{you2018graphrnn} sequentially evaluates each potential node and edge, requiring $O(N^2)$ inference passes to synthesize a graph with $N$ nodes. The Graph Recurrent Attention Network (GRAN)~\cite{liao2019gran} synthesizes all edges adjacent to a new node (or a block of nodes) in one shot, which reduces computational complexity but can also diminish the goodness of fit to the data. Since node embeddings were constructed by processing adjacency vectors using multilayer perceptrons (MLPs), GRAN also imposed a limit on the maximum graph size, corresponding to the size of the trained MLP.

In this paper, we present a new generative model for large graphs. At the core of our model is an \emph{auto-decoder}: a generative network that synthesizes a graph structure given a sequence of sampled vector codes (one code per node). During training, the parameters of the generator and the latent codes for the data are optimized jointly, end to end. This joint optimization is the key idea of the auto-decoder approach to generative modeling~\cite{tan1995reducing,bojanowski2018optimizing,park2019deepsdf}. We design a sequential decoder that uses graph attention modules in an autoregressive framework to synthesize coherent graph structures. By associating node embeddings to external codes, our auto-decoder model can synthesise graphs of any size, unconstrained by the maximum size of training graphs.

We observe and rectify a drawback in the auto-decoder framework: prior work optimized the latent codes during training, but then sampled codes at test time from an ad-hoc analytical distribution. This can induce a mismatch between training-time and test-time distributions in the latent space, reducing the fidelity of the synthesized structures. This can have a particularly significant effect in our setting, where latent codes are sampled many times during synthesis: one for each node in the synthesized graph. We rectify this deficiency by designing a flow-based density model~\cite{papamakarios2019normalizing}. This is a continuous model that learns to map random samples from an analytical distribution to samples that statistically match the empirical distribution of codes optimized by the auto-decoder. To capture relational structure among codes, our normalizing flow model is based on graph attention.

The resulting model convincingly outperforms the state of the art in generative modeling of large graph structures. Our model achieves the best rank across datasets according to all evaluation measures, while being 2x faster than the state of the art.

\section{Related Work}
\label{sec:Related Work}

{\bf Classic graph synthesis models.}
The Erd{\H{o}}s-R{\'e}nyi random graph model independently samples all the edges for generating a new graph~\citep{erdds1959random}. The Watts-Strogatz model generates graphs by randomly rewiring regular lattices~\citep{WattsStrogatz1998}. The Barab{\'a}si-Albert model~\citep{Barabasi1999} uses preferential attachment to synthesize graphs with power-law degree distributions. Exponential random graph models~\cite{wasserman1996logit} parametrize the distribution of real-world graphs using exponential families that use hand-engineered local structural features of graphs. The Kronecker graph model~\cite{leskovec2010kronecker} formulates the graph generation process as a recursion over Kronecker products of seed graphs. All these models have limited ability to fit real data and are known to not fully represent the structure of real-world networks~\cite{BroidoClauset2019,JacksonRogers2007,Lima-Mendez2009,you2018graphrnn}.

{\bf Tensor-based models.}
This class of generative models operates on tensor representations of graphs, such as adjacency matrices. This regular representation allows these models to generate a whole graph in one shot, akin to an image (a grid of pixels). Such models have been based on VAEs~\cite{Simonovsky2018, kipf2016variational, ma2018constrained}, GANs~\cite{de2018molgan}, and normalizing flows~\cite{liu2019graph, madhawa2019graphnvp}. They have been primarily demonstrated on generating small molecules and other small graphs. The one-shot synthesis of all nodes and edges assists parallel processing, but limits the maximal size of the generated graph and can yield incoherent structures.

{\bf Sequential generation models.}
These models formulate graph generation as a sequential decision process. Conditioned on the current subgraph structure, new nodes and edges are added in sequence~\cite{li2018learning}. Sequential models have the ability to examine intermediate structures and adapt subsequent synthesis to these. They are thus able to maintain accuracy for larger graph structures~\citep{you2018graphrnn,liao2019gran}. We build on and advance this line of work.

Domain-specific approaches have also been developed for molecular graphs. \citet{liu2018constrained} develop a sequential decoder that incorporates domain-specific constraints for generating valid molecular graphs. \citet{jin2018junction} utilize a sequential VAE and propose to operate on the junction tree of a molecular graph. \citet{you2018graph} train an adversarial network using reinforcement learning to optimize for molecules that satisfy desired properties. The autoregressive flow model of~\citet{shi2020graphaf} for molecular generation shares some similarity with our model in that it combines flow-based density estimation with a sequential decision process. However, unlike our model, the applicability of these works is limited to small domain-specific graphs.

{\bf Auto-decoders.} Auto-\underline{en}coder models have become widespread in machine learning. However, despite the name, the encoder is often discarded after training and the decoder is retained as a generative model. An elegant alternative is to not train with an encoder at all, but rather directly optimize the latent codes jointly with the generator by backpropagation. This idea goes back decades~\citep{tan1995reducing}, has been independently developed in a number of contexts~\citep{fan2018matrix}, and was recently popularized by~\citet{bojanowski2018optimizing}. We borrow the `auto-decoder' terminology from~\citet{park2019deepsdf}. We extend the auto-decoder framework to sequential generation of combinatorial structures, in which the generator operates on a sequence of sampled codes. We also rectify a major deficiency in the framework by augmenting it with a density model that is trained to sample codes from the learned distribution in the latent space.

{\bf Normalizing flows.}
Our density model is based on normalizing flows~\cite{papamakarios2019normalizing}. Flow-based models apply chains of invertible transformations to map samples from a latent space to data space and vice versa. They have been applied to variety of tasks, such as density estimation~\citep{dinh2014nice,germain2015made,45819}, variational inference~\citep{papamakarios2017masked,kingma2016improved}, image generation~\citep{kingma2018glow}, and audio synthesis~\citep{Prenger2019}. \citet{liu2019graph} developed graph normalizing flows to model latent node embeddings in a VAE framework. Flow-based models were also used to generate molecular graphs~\citep{shi2020graphaf,madhawa2019graphnvp}; these share the limitations of aforementioned models for molecules in that they do not scale to large graph structures.

{\bf Self-attention networks.}
Self-attention models dominate natural language processing~\citep{vaswani2017attention,Devlin2018,dai2019transformer,Yang2019xlnet} and have been applied to images~\citep{bello2019attention,ramachandran2019stand,brock2019large,Zhao2020} and combinatorial structures~\citep{velickovic2018graph,liao2019gran}. Our work uses self-attention modules due to their effectiveness in modeling global structure.

\section{Overview}
\label{sec:overview}

{\bf Notation.}
We aim to model an underlying distribution of the structure of given graph data $p(\GG)$ using a deep generative model. A graph $\G \in \GG$ is defined as $\G = (\V, \E)$, where $\V = \{v_1, \dots, v_N\}$ and $\E = \{(v_i, v_j) | i,j \in \{1, \dots, N\}\}$ represent the node set and connectivity structure, respectively. A graph can also be represented by its adjacency matrix $\AA \in \{0,1\}^{N\times N}$, where $A_{ij} = 1$ if and only if there exists an edge $(v_i, v_j)$ in $\E$. Given a node ordering $\pi$, there exists a correspondence between the set of permuted adjacency matrices $\AA^{\pi}$ and the set of node-ordered graphs $(\G, \pi)$.  For simplicity, we assume only undirected graphs with up to one edge between a pair of nodes and no self-loops. Thus due to the symmetry of the adjacency matrix, we can decompose $\AA^\pi = \LL^\pi + {\LL^\pi}^T$, where $\LL^\pi$ denotes the corresponding lower-triangular component of $\AA^\pi$ and $\cdot^T$ denotes the transpose. Consequently, we can model $p(\G) = \sum_{\pi} p(\G, \pi) = \sum_{\pi} p(\AA^\pi) = \sum_{\pi} p(\LL^\pi)$ by modeling the distribution over lower-triangular matrices $\LL^\pi$.

{\bf Model.}
We refer to our model as a Graph Auto-Decoder (GrAD). The training has two stages. The first stage trains a sequential auto-decoder $\D_\btheta$ that learns to generate the matrix $\LL^{\pi}$ while simultaneously optimizing the latent codes $\ZZ = \{\zz_i\}_{i=1}^N \in \Re^{N\times d}$ that serve as input to $\D_\btheta$. In the second stage, we train a flow-based density model $\F_{\bm\Phi}$ that learns to map samples from a simple analytical distribution (a Gaussian) to the distribution of latent codes $\ZZ$ optimized in the first stage. $\btheta$ and $\bm\Phi$ are the parameters of the auto-decoder and the density model, respectively. Note that GrAD uses \emph{per-node} latent codes. The batch of codes $\ZZ$ correspond to a single graph. $\ZZ$ contains a code $\zz_i$ for each node in this graph.

{\bf Inference.}
The number of nodes in a new graph is sampled from an empirical distribution: $N \sim p(N)$. The density model $\F_{\bm\Phi}$ is then used to map a batch of $N$ vectors sampled from a Gaussian distribution ($\YY \sim \N(0, I_{N\times d})$) into $N$ codes of the same dimensionality but distributed according to the learned distribution of latent codes: $\ZZ = \F_{\bm\Phi}^{-1}(\YY)$. The decoder then sequentially synthesizes the graph by appending blocks of nodes. Given a new block, a scaffold of putative edges is constructed within this block and from the block to the previously synthesized subgraph. Latent codes for the nodes are then processed via graph self-attention layers that operate on this scaffolded structure. The transformed features are then used to synthesize parameters for probability distributions for each putative edge. The existence of each edge is then sampled from the corresponding synthesized distribution.

\begin{figure*}[ht]
\centering
\includegraphics[width=0.9\textwidth]{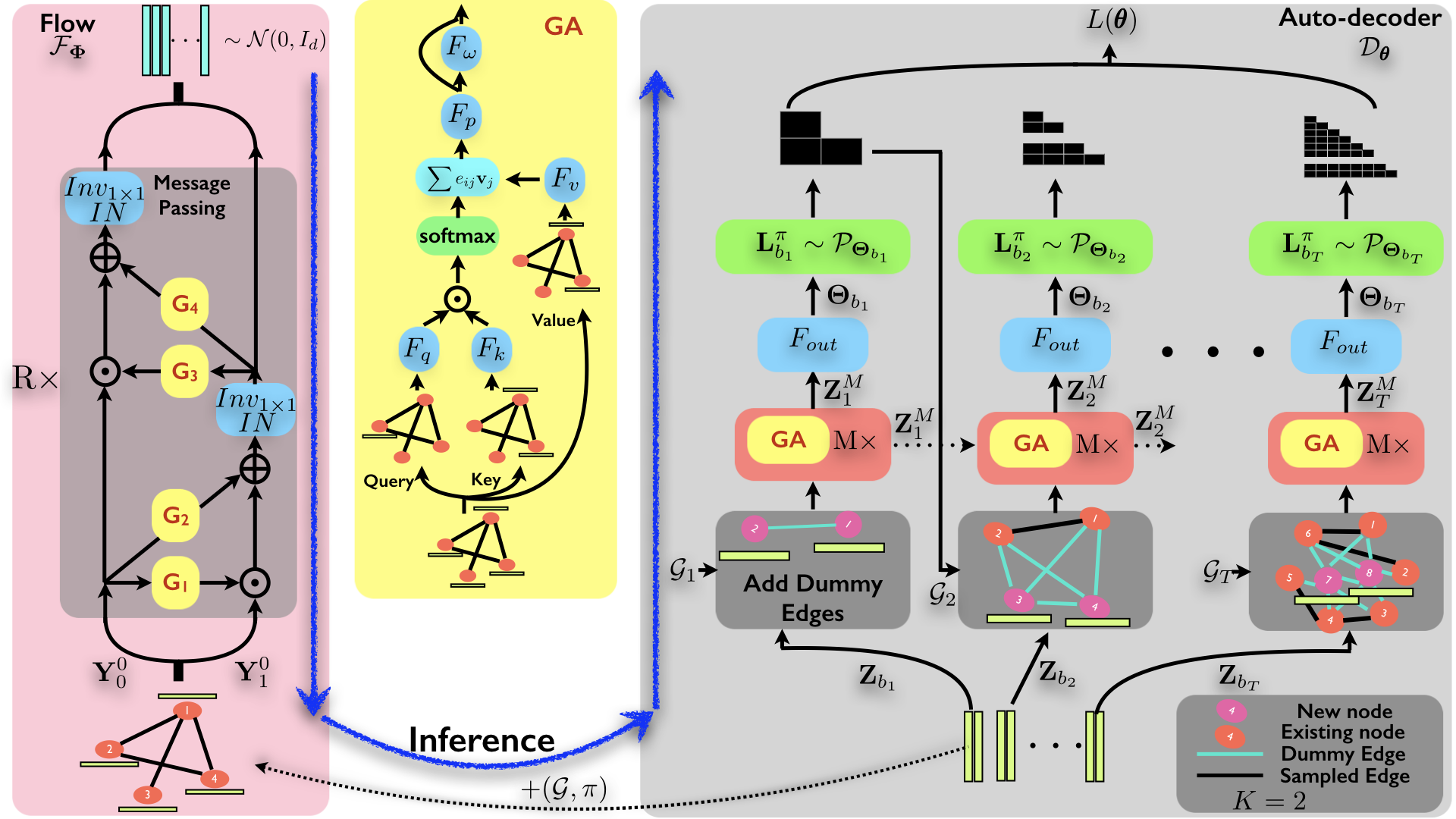}
\caption{GrAD trains in two stages. In the first stage, the auto-decoder $\D_\btheta$ (right) is trained on a set of observed graphs. The training of $\D_\btheta$ jointly optimizes its parameters along with the latent codes that serve as its input. In the second stage, a flow-based density model $\F_{\Phi}$ (left) is trained to map from a Gaussian distribution to the distribution of optimized latent codes. Both models use graph attention (GA) layers, visualized in the middle.}
\label{fig:model}
\vspace{-3mm}
\end{figure*}

\section{Graph Auto-decoder}
\label{sec:model}

The auto-decoder $\D_\btheta$ formalizes the problem of learning a distribution over matrices $\LL^\pi$ as a sequential decision process over nodes. However, in order to efficiently scale inference over large graphs, GrAD does not generate edges sequentially~\cite{you2018graphrnn}. Instead, we generate all entries of each row (or a block of rows) of the adjacency matrix jointly, conditioned on the previously estimated submatrix~\cite{liao2019gran}. At each time step $t$, $\D_\btheta$ generates $K$ nodes and their edge connectivity with respect to all existing nodes. Equivalently, it samples a block of $K$ rows $\LL^\pi_{b_t} = [\LL^\pi_{K(t-1) + 1}, \dots, \LL^\pi_{Kt}]$ of $\LL^\pi$, where a vector $\LL^\pi_i \in \Re^{1\times N}$ denotes the $i$th row and $b_t = [K(t-1) + 1, \dots, Kt]$ is a set of indices.

Figure~\ref{fig:model} (right) illustrates the procedure in detail. Starting from an empty graph $\G_1$, in each step our model samples the latent variables for a new block of nodes $\ZZ_{b_t}$ and processes them alongside the existing subgraph $\{\LL^\pi_{b_i}\}_{i=1}^{t-1}$ and its hidden node representation $\ZZ_{t-1}^M$ (inherited from the previous time step). Afterwards, the connectivity structure $\LL^\pi_{b_t}$ within this new set of nodes and across to the previously synthesized nodes is generated in a single shot according to the corresponding latent codes $\ZZ_t^M$. Overall, $\D_\btheta$ iterates over $T=\ceil{\frac{N}{K}}$ time steps until all the nodes and their edges are generated. This sequential generation process decomposes the probability distribution of $\LL^\pi$ into a product of block conditional distributions~\cite{liao2019gran}:
\begin{align}
p(\LL^\pi | \ZZ) &= \prod_{t=1}^T p(\LL^\pi_{b_t} | \LL^\pi_{b_{t-1}}, \dots, \LL^\pi_{b_1}, \ZZ).
\end{align}
The auto-decoder applies multiple layers of graph self-attention. Specifically, at each time step $t$, the decoder takes a set of node features $\ZZ_t^0$ and a partial adjacency matrix $\{\LL^\pi_{b_i}\}_{i=1}^{t-1}$ as input. It then constructs a subgraph $\G_t$ that contains all existing and new nodes, with previously synthesized edges between the prior nodes. Dummy (putative) edges that connect each new node with all other nodes in the graph $\G_t$ are added. This ensures that subsequent graph attention layers can propagate information to and from the new nodes. Using this scaffold, new node embeddings are computed by $M$ layers of multi-headed graph attention (GA). The model then estimates the parameters of a probability distribution for each edge, $\bm\Theta_{b_t}$, and samples edges from their probability distributions:
 \begin{align}
 \ZZ_t^l &= \text{GA}\paren{\ZZ_t^{l-1}; \{\LL^\pi_{b_i}\}_{i=1}^{t-1}}, \quad l=1,\dots,M \\
 \bm\Theta_{b_t} &= F_{out}\paren{\ZZ_t^M} \\
 \LL^\pi_{b_t} &\,\sim \P_{\bm\Theta_{b_t}}
 \end{align}
where $\ZZ_t^l \in \Re^{Kt\times d}$ is the hidden node representation at the $l$-th message passing layer and GA is the graph attention module. At each time step, the initial hidden node representation $\ZZ_t^0$ is set to $\ZZ_t^0 = [\ZZ_{t-1}^M;  \ZZ_{b_t}]$, where $[\cdot]$ stands for vector concatenation. Both GA and $F_{out}$ modules share weights across all time steps. However, the $M$ stacked GA layers have independent parameters.

\subsection{Graph Attention}
\label{sec:GA}

As discussed above, at each time step, node features are processed using a graph attention (GA) mechanism. This operation is implemented as a GA layer that is utilized throughout our model.
We now describe this layer. We denote the node features provided as input to GA by ${\ZZ \in \Re^{m\times d}}$.

GA begins by mapping $\ZZ$ to matrices of query, key, and value using linear transformations~\citep{vaswani2017attention}:
\begin{align}
\QQ = F_q(\ZZ; \btheta_Q); \,\, \KK = F_k(\ZZ; \btheta_K); \,\, \VV = F_v(\ZZ; \btheta_V), \label{eqn:gau1}
\end{align}
where $F_i$s are two-layer MLP mappings $F_i : \Re^d \to \Re^{d_S}$ and $d_S$ is the dimensionality of the transformed features. The self-attention weight of the $j$th node to node $i$ is computed as
\begin{align}
a_{ij} &= \text{softmax}(\ee_i)_j = \frac{\exp(e_{ij})}{\sum_{k\in\N_i} \exp(e_{ik})}, \quad \text{where} \,\, e_{ij} = {d_S}^{-\frac{1}{2}}\paren{\qq_i^T\kk_j}. \label{eqn:gau2}
\end{align}
Here $\qq_i$ denotes the $i$th row of matrix $\QQ$ and a set ${\N_i = \{k | (v_i, v_k) \in \E_{\G_t}\}}$ represents the 1-ring neighborhood of node $i$ in subgraph $\G_t$. By restricting each node's attention to its neighbors $\N_i$, the model provides structural information from the current subgraph $\G_t$. Updated features are composed by a linear combination of value vectors:
\vspace{-4mm}
\begin{align}
\tilde{\zz}_i &= \sum_{j\in\N_i} a_{ij} \vv_j.
\label{eqn:gau4}
\end{align}
GA employs multiple heads~\citep{vaswani2017attention}, each independently transforming the input features $\ZZ$ using the attention mechanism summarized in \mbox{Equations~(\ref{eqn:gau1}-\ref{eqn:gau4})}. This is followed by concatenating the output features from different heads and projecting back to the input feature dimensionality $d$ using a linear operator $\btheta_P \in \Re^{Hd_S \times d}$:
\vspace{-2mm}
\begin{align}
\tilde{\ZZ} &= F_p\paren{[\tilde{\ZZ}^h]_{h=1}^H; \btheta_p},
\end{align}
where $\tilde{\ZZ}^h$ is the output of self-attention head $h$. The features are then normalized using layer normalization (LN)~\cite{ba2016layer} and processed by a two-layer MLP with ReLU nonlinearities, $F_{\omega}$:
\begin{align}
\ZZ' &= \text{LN}\paren{\hat{\ZZ} + F_{\omega}\paren{\hat{\ZZ}; \btheta_\omega}} \text{, where} \,\, \hat{\ZZ} = \text{LN}\paren{\ZZ + \tilde{\ZZ}}\label{eqn:fphi}
\end{align}

\subsection{Sampling Edges}

After $M$ rounds of processing via GA layers, the resulting node features $\ZZ_t^M$ are processed by a block $F_{out}$ that estimates the sampling distribution parameters $\bm\Theta_{b_t}$ for each individual edge in $\LL^\pi_{b_t}$.
$\D_\theta$ models edge probability in block $\LL^\pi_{b_t}$ as a finite mixture of multivariate Bernoulli distributions~\cite{dai2013multivariate} parametrized by $\bm\Theta_{b_t}=[\{\Pi^c\}_{c=1}^C, \{\bm\lambda^c\}_{c=1}^C]$. Here $\Pi^c$ represents the mixing proportion and $C$ is the total number of mixture components. In particular, the function $F_{out}$ implements two three-layer MLPs with ReLU nonlinearities and $C$-dimensional output, $F_\lambda$ and $F_\Pi$, which independently compute $\bm\lambda_{ij}$ and $\bm\Pi_{ij}$ from $\ZZ_t^M$.
\begin{align}
p(\LL^\pi_{b_t} | \LL^\pi_{b_<{t}}, \ZZ; \btheta) = \displaystyle \sum_{c=1}^C \Pi^c \,\,p(\LL^\pi_{b_t} | \bm\lambda^c)
= \displaystyle \sum_{c=1}^C \Pi^c &\prod_{i\in \bb_t} \prod_{j < i} [\lambda^c_{ij}]^{\epsilon_{ij}} [1 - \lambda^c_{ij}]^{1 - \epsilon_{ij}} \label{eqn:out} \\
\Pi^1, \dots, \Pi^c = \text{softmax}\paren{\displaystyle\sum_{i\in \bb_t} \sum_{j<i} F_\Pi\paren{\zz_i - \zz_j; \btheta_\Pi}};  \quad
&\lambda^1_{ij}, \dots, \lambda^c_{ij} = \sigma\paren{F_{\lambda}\paren{\zz_i - \zz_j; \btheta_\lambda}} \nonumber
\end{align}
where $\epsilon_{ij} = 1 \text{ if and only if } (v_i, v_j)\in \E$, and $\sigma$ denotes the sigmoid activation function. Note that the edge distributions are independent within a component but not across components. Further, the distribution of $\bm\Pi$ in Equation~\eqref{eqn:out} exchanges global information. Thus unlike a single Bernoulli distribution (i.e., $C=1$), a mixture can capture correlations between edges. In the absence of sequential edge generation, employing a finite mixture allows our model to encode complex dependencies across edges in the output distribution of $\LL^\pi_{b_t}$.

\section{Optimization}
\label{sec:opt}

The graph generation process in the decoder can be specified as $\ZZ \sim p(\ZZ)$ followed by ${\LL^\pi \sim p(\LL^\pi | \ZZ; \btheta)}$. The joint distribution can be written as ${p(\LL^\pi, \ZZ; \btheta) = p(\ZZ)p(\LL^\pi | \ZZ; \btheta)}$. Given an observed set of matrices $\{\LL^\pi_i\}_{i=1}^n$ sampled from an unknown distribution $p(\LL^{\pi})$, the auto-decoder $\D_\btheta$ is trained by maximizing the marginal log-likelihood of the observations while integrating out the latent variables:
\begin{align}
L(\btheta) = \displaystyle \frac{1}{n}\sum_{i=1}^n \log p(\LL^\pi_i;  \btheta) = \displaystyle \frac{1}{n}\sum_{i=1}^n \log \int_\ZZ p(\LL^\pi_i, \ZZ; \btheta) d\ZZ \label{opt:ll}
\end{align}
The common practice is to use variational expectation-maximization (vEM)~\cite{bishop2006pattern} to iteratively learn model parameters $\btheta$ and posterior distribution $p\paren{\ZZ | \LL^\pi_i; \btheta}$ for $\ZZ$.
An alternative is to optimize a differentiable model in~\eqref{opt:ll} using SGD with a constant learning rate, which also leads to simple vEM algorithm~\citep{mandt2017stochastic}. Thus, given an observed matrix $\LL^\pi_i$, GrAD jointly learns the model parameters $\btheta = \{\btheta_Q, \btheta_K, \btheta_V, \btheta_P, \btheta_\omega, \btheta_\Pi, \btheta_\lambda\}$ and latent variables $\ZZ_i$'s using the following update rule:
\begin{equation}
\resizebox{.9\hsize}{!}{$
\btheta^{T+1} = \btheta^{T} + \tau\left[\frac{1}{n}\sum_{i=1}^n  \nabla_\btheta \log p\left(\LL^\pi_i, \ZZ_i^T; \btheta^T\right)\right] \quad
\ZZ_i^{T+1} = \ZZ_i^T + \delta \left[\nabla_\ZZ \log p\left(\LL^\pi_i, \ZZ_i^T; \btheta^T\right)\right] \label{opt:sgd}
$}\\
\end{equation}
where step size $\delta$ is kept fixed. After each update, the latent codes are projected back onto the unit $\ell_\infty$ ball~\cite{bojanowski2018optimizing}. We initialize $\ZZ$ by sampling from the prior distribution ${\ZZ \sim p(\ZZ) = \N(0, I_{N\times d})}$. The gradient computation for both updates in~\eqref{opt:sgd} shares the same chain rule.

\section{Density Model}

At inference time, we must produce latent codes $\ZZ$ that are provided as input to $\D_{\btheta}$. One possibility is to draw them from an analytical distribution, such as a Gaussian. However, this produces uncorrelated sets of codes whose joint distribution is very different from the joint distribution of code sets that are optimized by $\D_{\btheta}$ during training. To rectify this, we train a flow-based reversible model to map the optimized code sets to a Gaussian. The inverse of this model can then be used to map Gaussian samples to the latent distribution of the optimized code sets.

The flow $\F_{\bm\Phi}$ processes a set of codes $\ZZ$ with a corresponding connectivity structure $\G$ via $R$ reversible message passing steps. Invertibility is achieved by splitting the dimensions into two parts, $\YY^0 \doteq \ZZ = \left[\YY^0_0, \YY^0_1\right]$, and operating on each part in turn~\cite{dinh2014nice}. The $l$th message passing step has the following structure:
\begin{align}
\YY^{l + \frac{1}{2}}_0 = \YY^{l}_0 &\quad\quad
\YY^{l + \frac{1}{2}}_1 = N_1\paren{\YY^{l}_1 \odot \exp\paren{G_1\paren{\YY^{l}_0}} + G_2\paren{\YY^{l}_0}}. \\
\YY^{l + 1}_1 = \YY^{l + \frac{1}{2}}_1 &\quad\quad
\YY^{l + 1}_0 = N_2\paren{\YY^{l + \frac{1}{2}}_0 \odot \exp\paren{G_3\paren{\YY^{l+ \frac{1}{2}}_1}} + G_4\paren{\YY^{l+ \frac{1}{2}}_1}}.
\end{align}
Here $N_i = \text{Inv}_{1\times 1} \circ \text{IN}$ represents instance normalization followed by a learnable channel permutation using an invertible $1\times 1$ convolution. Each $G_i$ denotes a GA layer (Section~\ref{sec:GA}) operating on the graph $\G$. $\F_{\bm\Phi}$ is trained by maximizing the exact log-likelihood of latent node embeddings. Using the change-of-variables formula,
\begin{align}
\log p(\ZZ; \Phi) &= \log p(\YY^R) + \displaystyle \sum_{l=1}^R \log \paren{\det \abs{\frac{\partial \YY^l}{\partial \YY^{l-1}}}},
\end{align}
where $p(\YY^R) = \prod_{j=1}^N \N(\yy^R_j| 0; I_d)$ and $\Phi$ are the model parameters. The Jacobian of $\YY^l$ at $\YY^{l-1}$, $\frac{\partial \YY^l}{\partial \YY^{l-1}}$, is a lower-triangular matrix that can be computed efficiently.

At inference time, $\F_{\bm\Phi}$ transforms a fully-connected graph on a set of Gaussian samples through the inverse of the learned flow.

\begin{figure*}[!htb]
\centering
\ra{1.05}
\resizebox{1\linewidth}{!}{
\begin{tabular}{@{}l@{\hspace{1mm}}c@{\hspace{2mm}}c@{\hspace{2mm}}c@{\hspace{2mm}}|c@{\hspace{2mm}}c@{\hspace{2mm}}c@{\hspace{2mm}}|c@{\hspace{2mm}}c@{\hspace{2mm}}c@{}}
& \multicolumn{3}{c}{Lobster} & \multicolumn{3}{c}{Community} & \multicolumn{3}{c}{Protein} \\
\rotatebox[origin=l]{90}{Train} &
\includegraphics[width=0.11\linewidth]{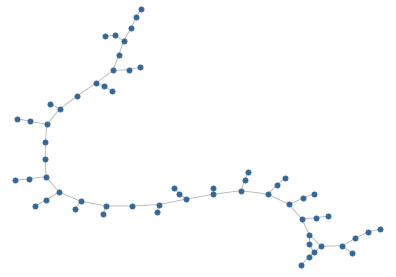} &
\includegraphics[width=0.11\linewidth]{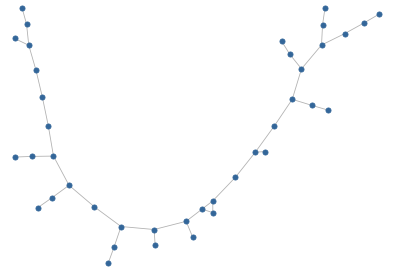} &
\includegraphics[width=0.11\linewidth]{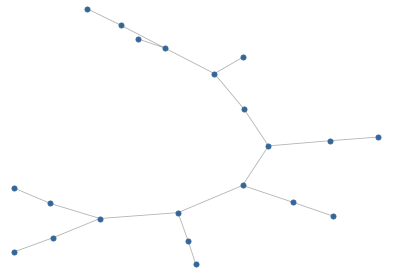} &
\includegraphics[width=0.11\linewidth]{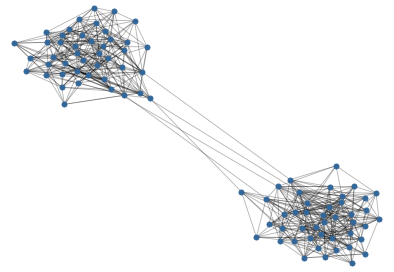} &
\includegraphics[width=0.11\linewidth]{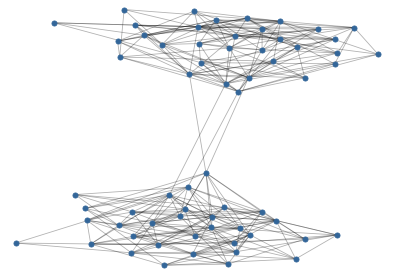} &
\includegraphics[width=0.11\linewidth]{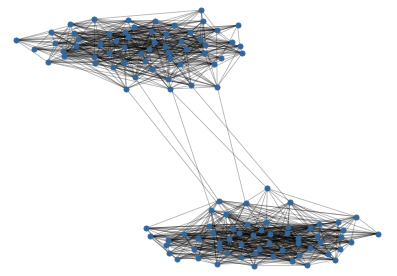} &
\includegraphics[width=0.11\linewidth]{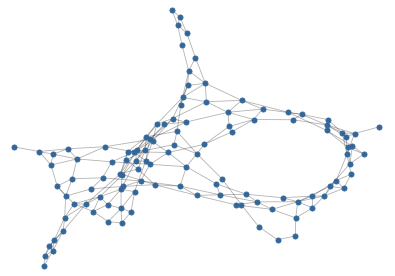} &
\includegraphics[width=0.11\linewidth]{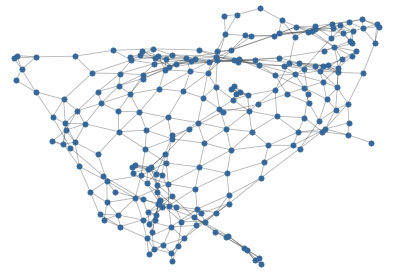} &
\includegraphics[width=0.11\linewidth]{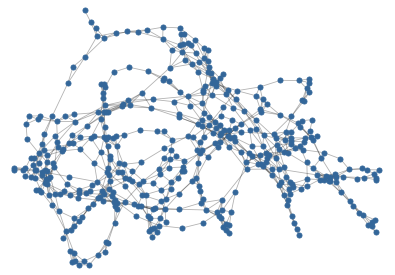}
\\
\rotatebox[origin=l]{90}{G-RNN} &
\includegraphics[width=0.11\linewidth]{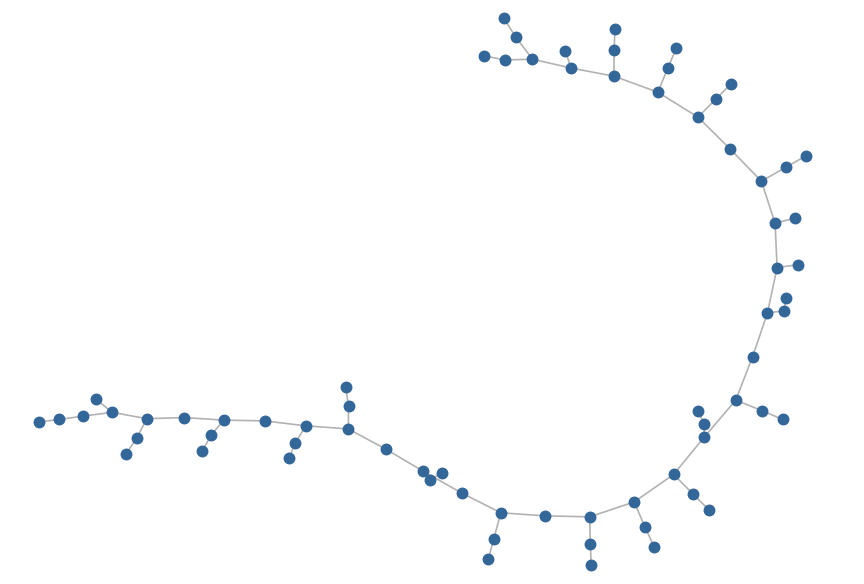} &
\includegraphics[width=0.11\linewidth]{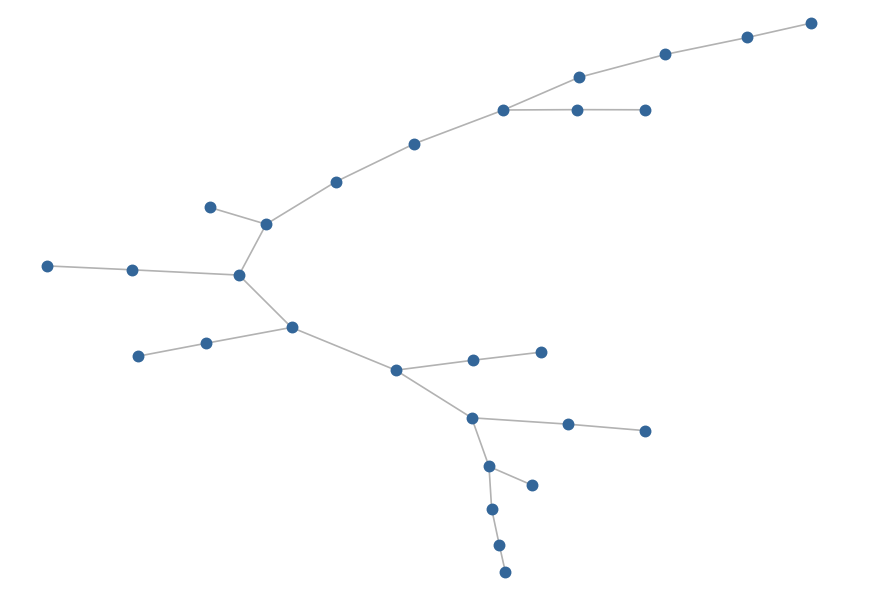} &
\includegraphics[width=0.11\linewidth]{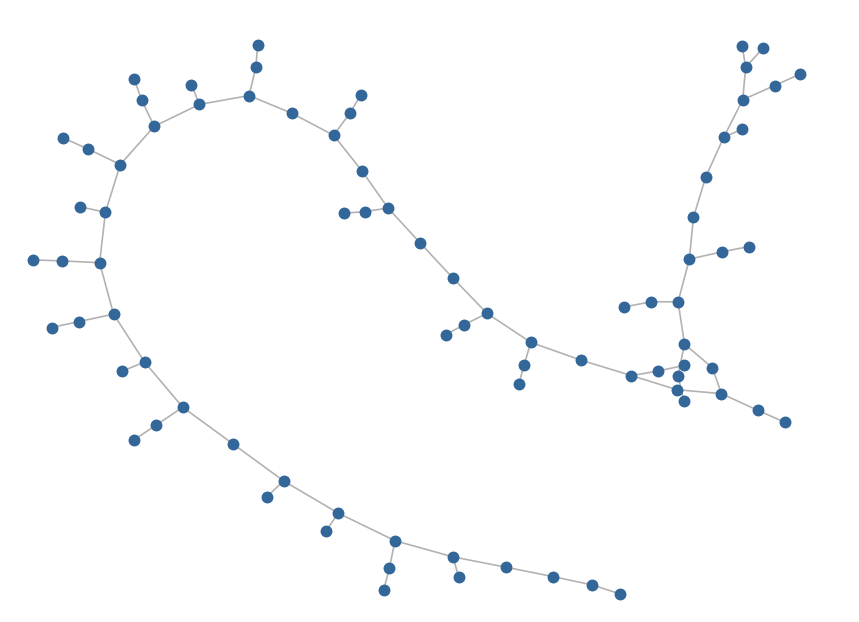} &
\includegraphics[width=0.11\linewidth]{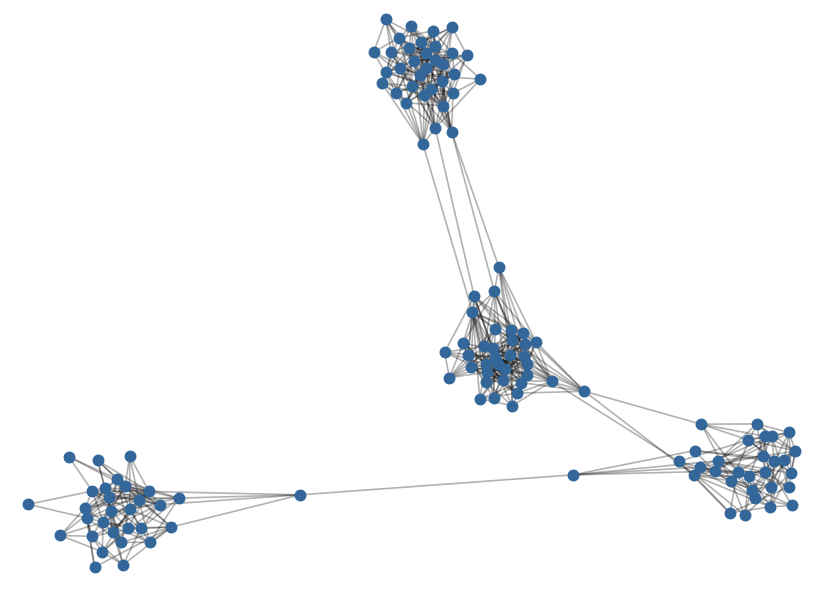} &
\includegraphics[width=0.11\linewidth]{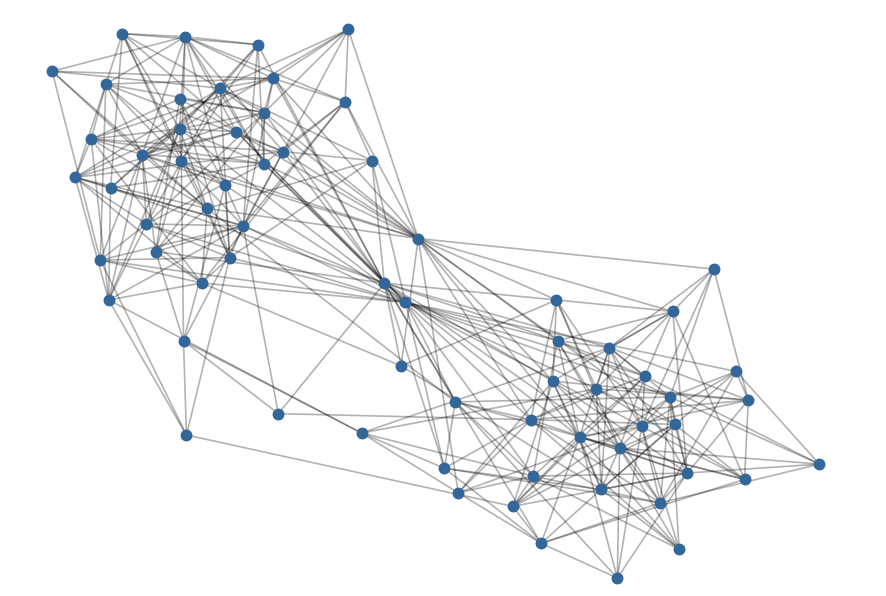} &
\includegraphics[width=0.11\linewidth]{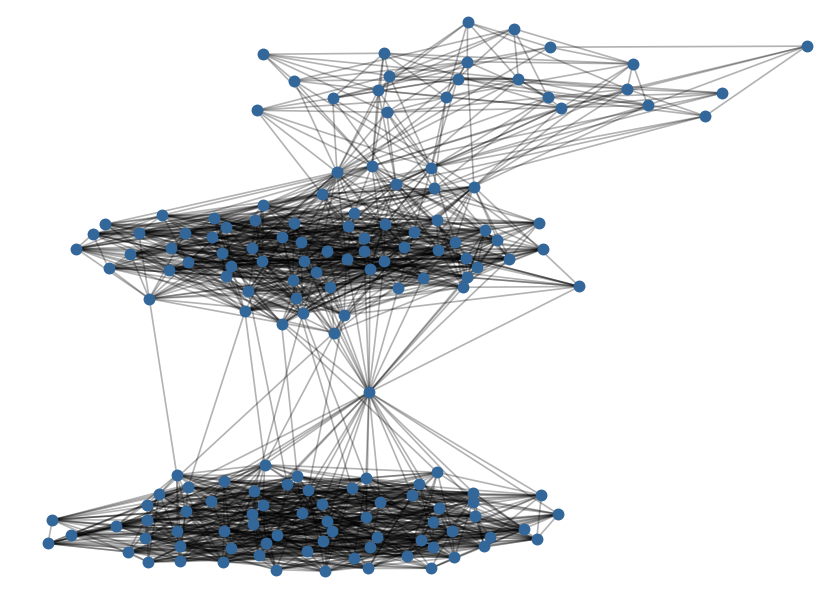} &
\includegraphics[width=0.11\linewidth]{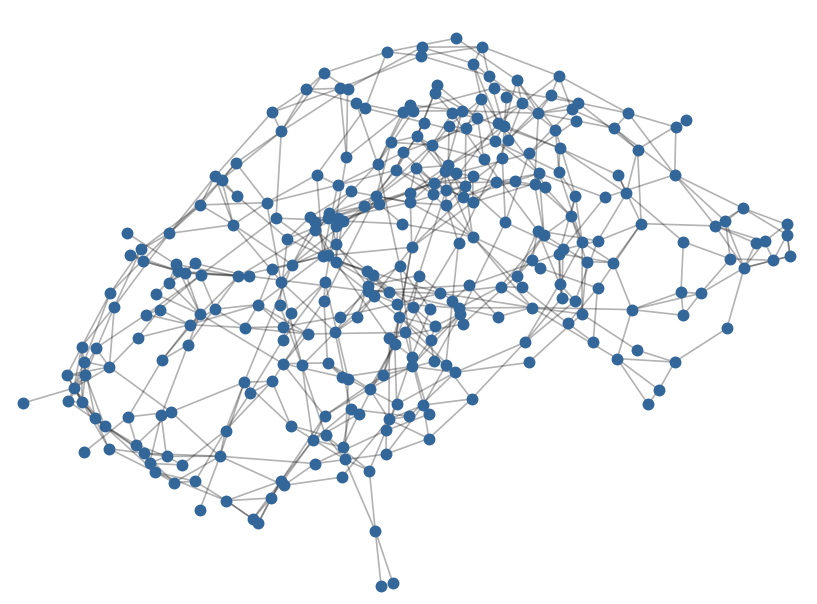} &
\includegraphics[width=0.11\linewidth]{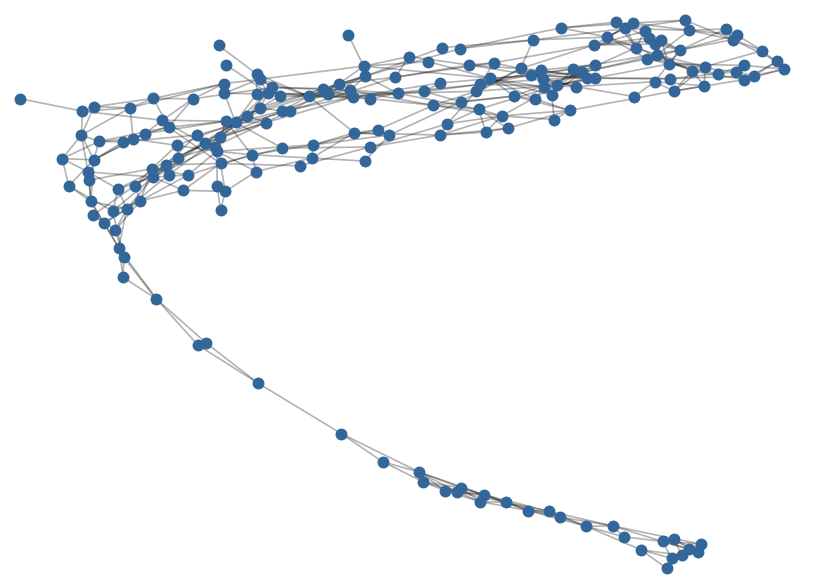} &
\includegraphics[width=0.11\linewidth]{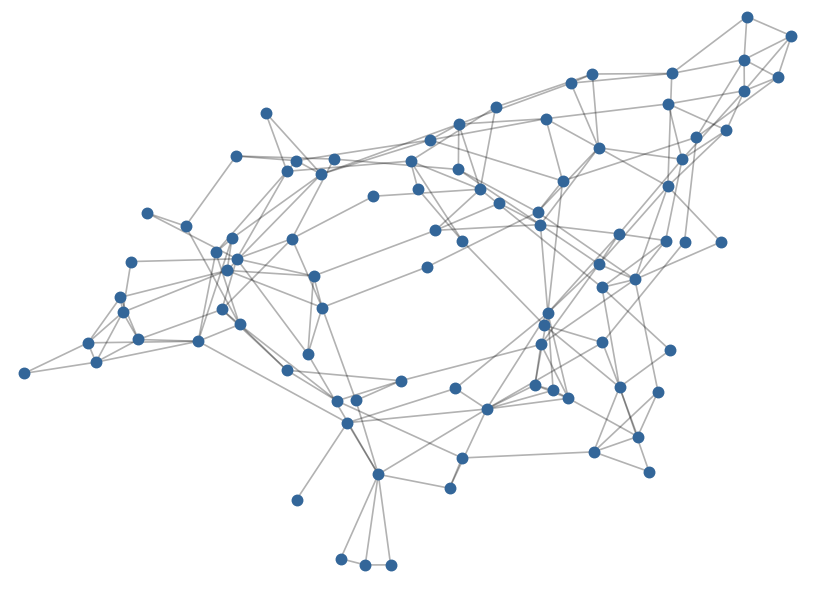}
\\
\rotatebox[origin=l]{90}{GRAN} &
\includegraphics[width=0.11\linewidth]{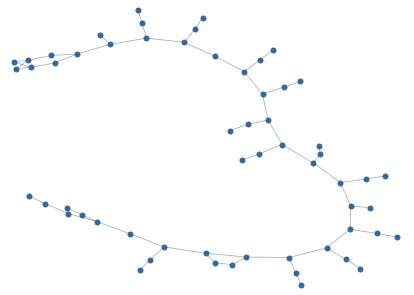} &
\includegraphics[width=0.11\linewidth]{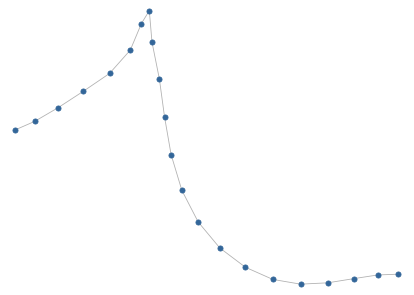} &
\includegraphics[width=0.11\linewidth]{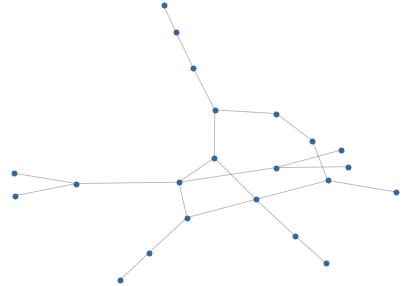} &
\includegraphics[width=0.11\linewidth]{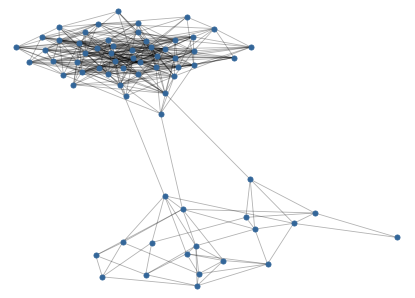} &
\includegraphics[width=0.11\linewidth]{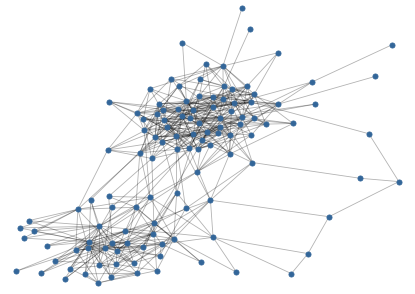} &
\includegraphics[width=0.11\linewidth]{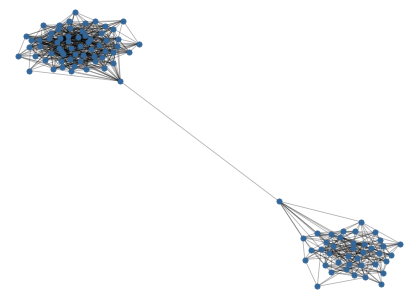} &
\includegraphics[width=0.11\linewidth]{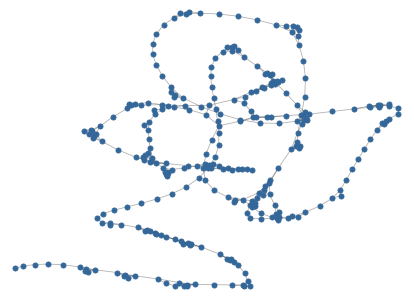} &
\includegraphics[width=0.11\linewidth]{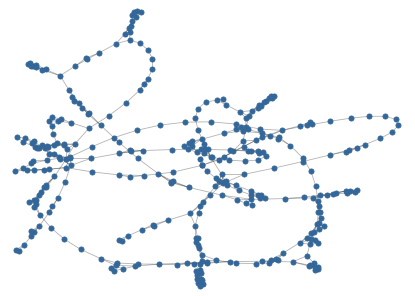} &
\includegraphics[width=0.11\linewidth]{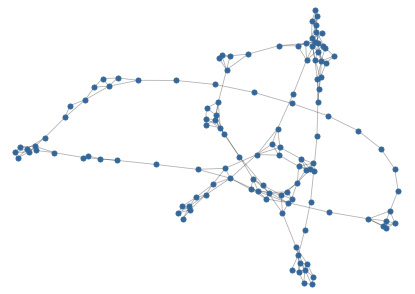}
\\
\rotatebox[origin=l]{90}{GrAD} &
\includegraphics[width=0.11\linewidth]{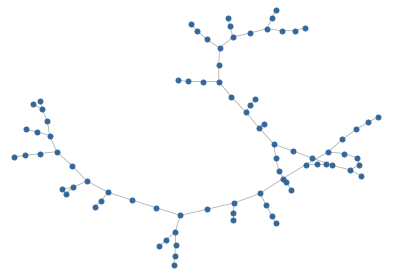} &
\includegraphics[width=0.11\linewidth]{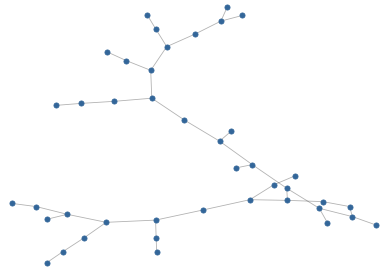} &
\includegraphics[width=0.11\linewidth]{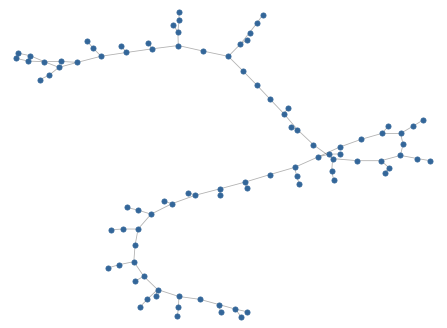} &
\includegraphics[width=0.11\linewidth]{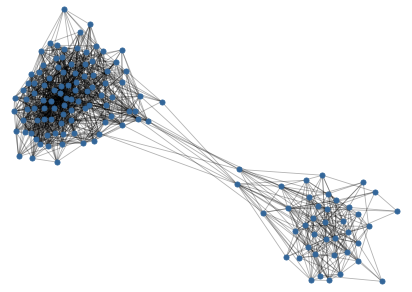} &
\includegraphics[width=0.11\linewidth]{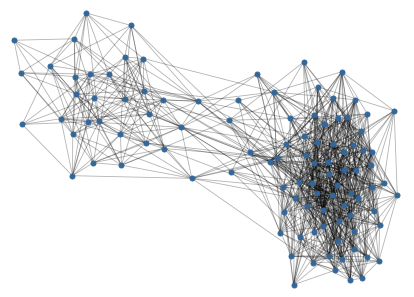} &
\includegraphics[width=0.11\linewidth]{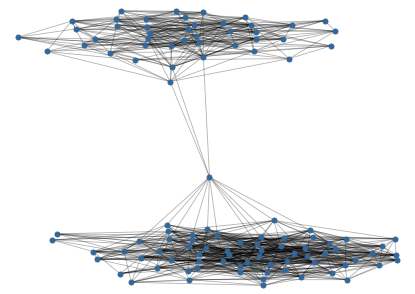} &
\includegraphics[width=0.11\linewidth]{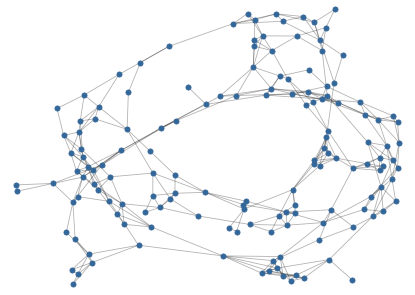} &
\includegraphics[width=0.11\linewidth]{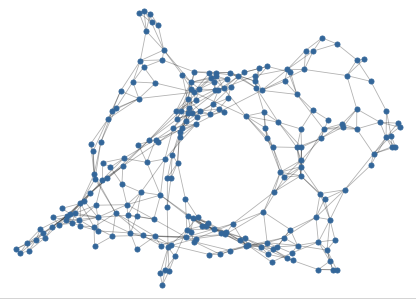} &
\includegraphics[width=0.11\linewidth]{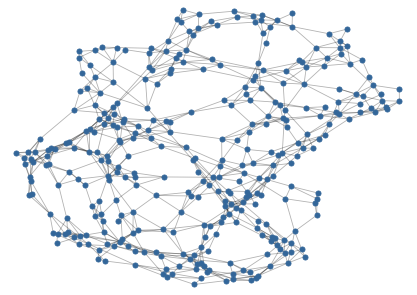}
\\
\rotatebox[origin=l]{90}{Unseen} &
\includegraphics[width=0.11\linewidth]{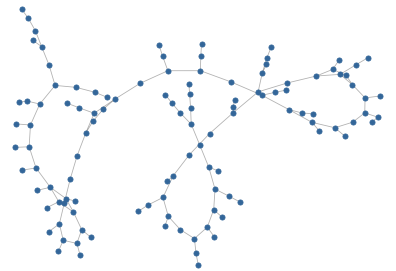} &
\includegraphics[width=0.11\linewidth]{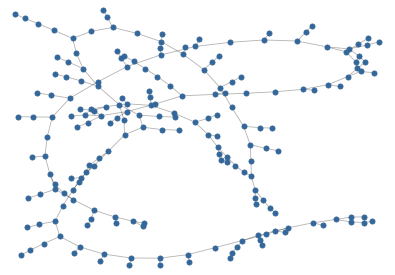} &
\includegraphics[width=0.11\linewidth]{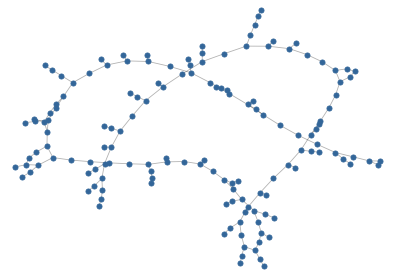} &
\includegraphics[width=0.11\linewidth]{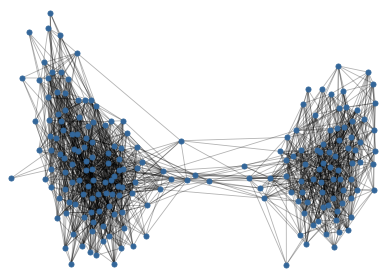} &
\includegraphics[width=0.11\linewidth]{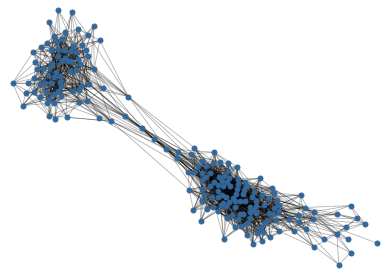} &
\includegraphics[width=0.11\linewidth]{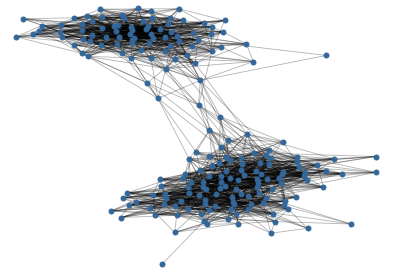} &
\includegraphics[width=0.11\linewidth]{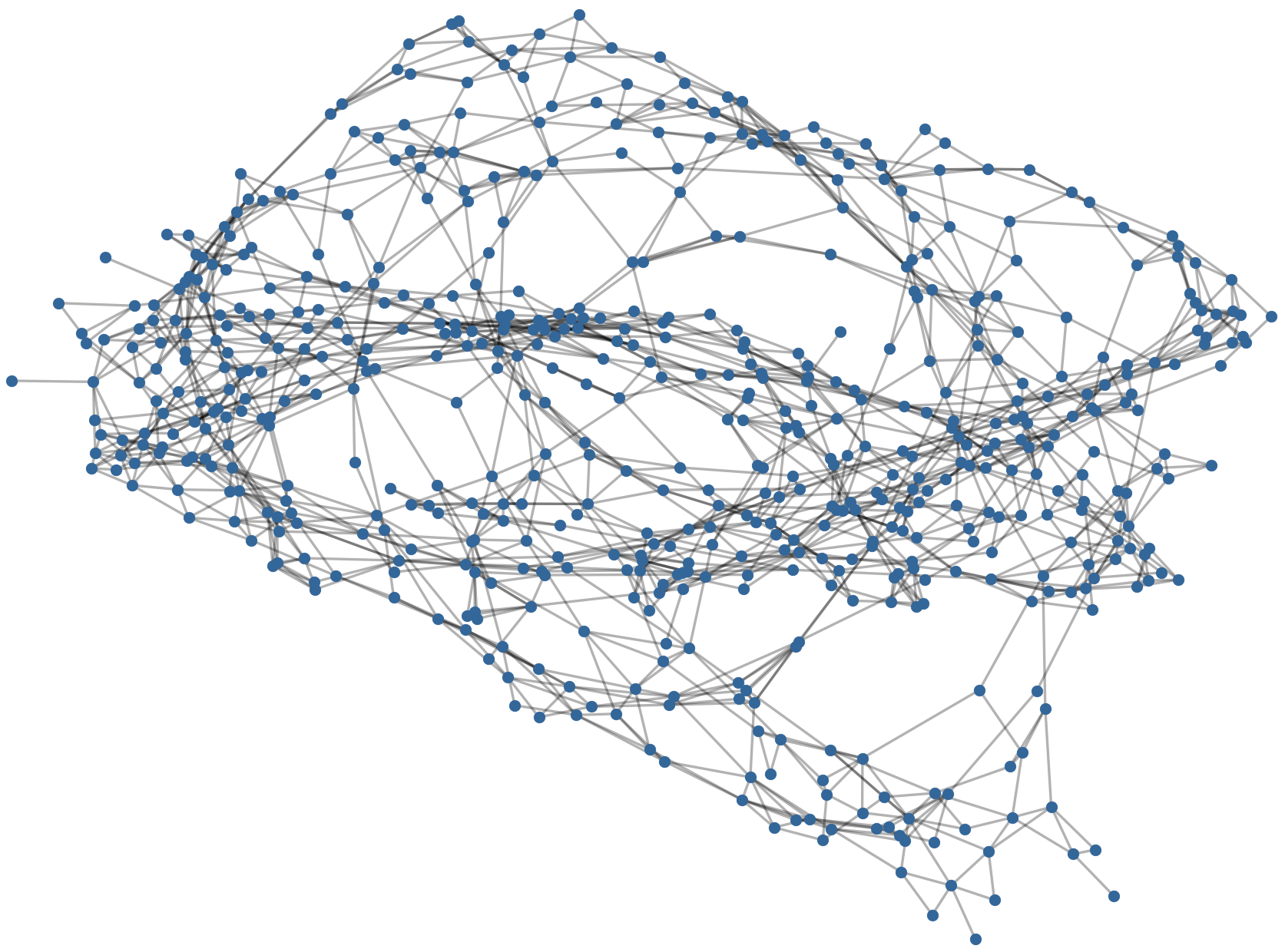} &
\includegraphics[width=0.11\linewidth]{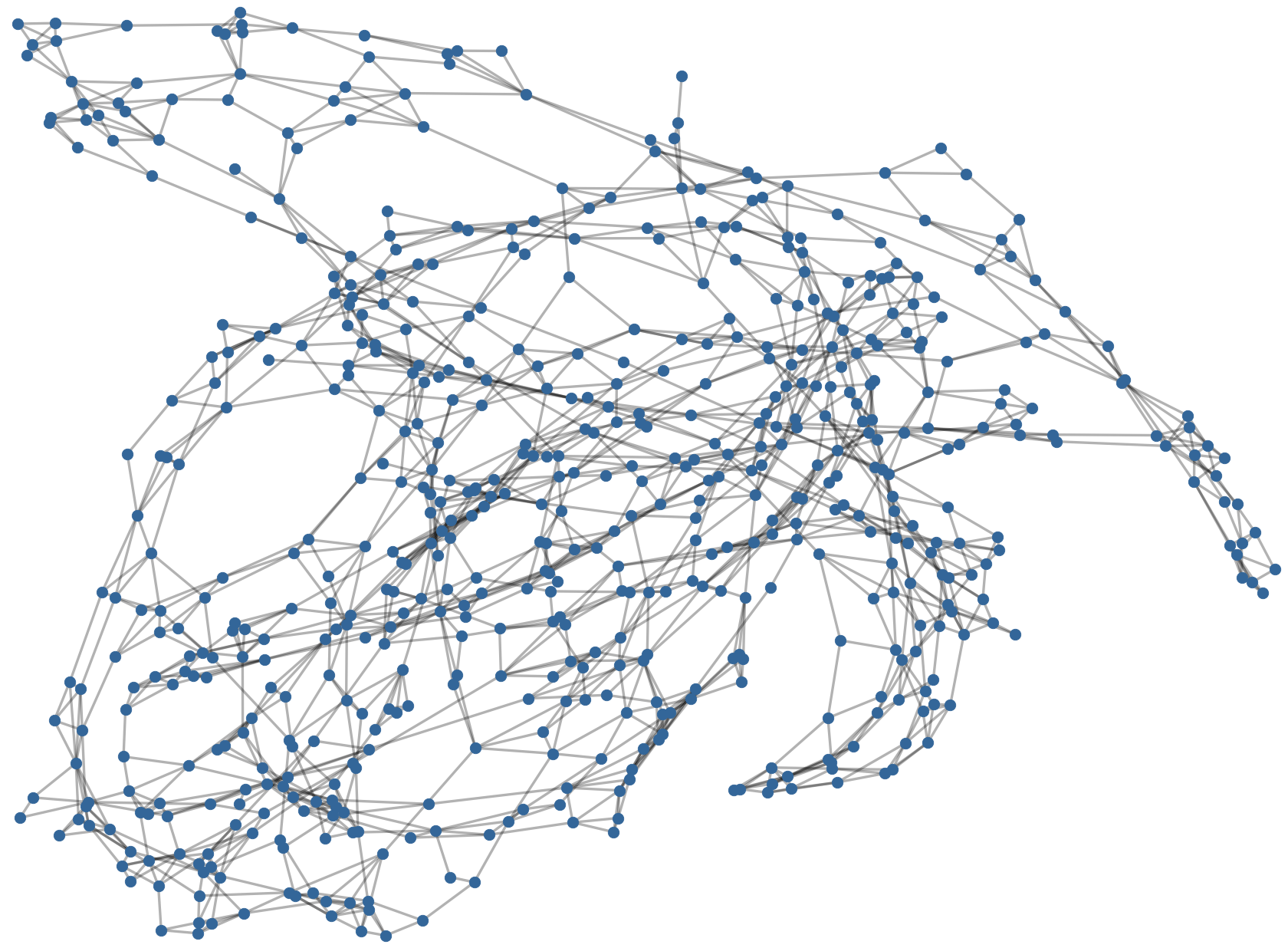} &
\includegraphics[width=0.11\linewidth]{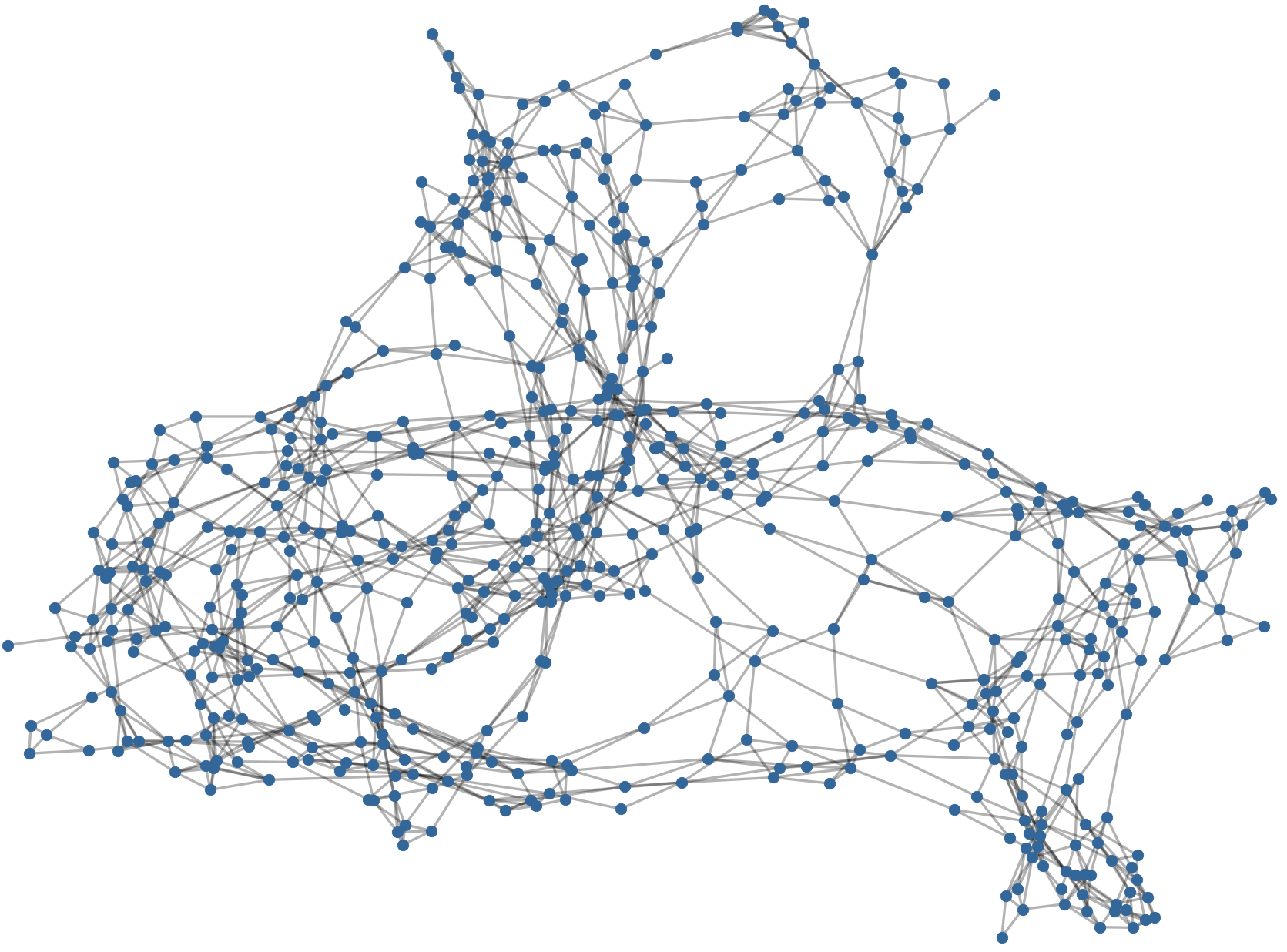}
\end{tabular}
}
\caption{Random samples of graphs generated by GraphRNN, GRAN, and GrAD (our approach). The bottom row shows graphs generated by GrAD for cardinalities that are beyond those observed in the training data.}
\label{fig:qual}
\end{figure*}

\begin{table*}[!htb]
\centering
\ra{1.05}
\resizebox{1\linewidth}{!}{
\begin{tabular}{@{}l@{\hspace{1mm}}|c@{\hspace{2mm}}c@{\hspace{2mm}}c@{\hspace{2mm}}c@{\hspace{2mm}}c@{\hspace{2mm}}c|@{\hspace{2mm}}c@{\hspace{2mm}}c||@{\hspace{1mm}}c@{\hspace{2mm}}c@{\hspace{2mm}}c@{\hspace{2mm}}c@{\hspace{2mm}}c@{\hspace{2mm}}c|@{\hspace{2mm}}c@{\hspace{2mm}}c@{}}
\toprule
Algorithm & Cycle & Grid & Lobs. & Comm. & Ego & Prot. & Mean & Rank & Cycle & Grid & Lobs. & Comm. & Ego & Prot. & Mean & Rank\\
\midrule
 & \multicolumn{8}{c}{Degree} & \multicolumn{8}{c}{Orbit} \\
\midrule
GraphRNN & 0.013 & 0.007 & $\pmb{\bm2e^{-4}}$ & 0.038 & 0.047 & 0.005 & 0.018 & 3.2 & $2e^{-4}$ & 0.007 & $1.1e^{-5}$ & \textbf{0.015} & 0.260 & 0.370 & 0.109 & 2.8 \\
GraphRNN-S & 0.006 & 0.041 &  0.005 & 0.130 & 0.046 & 0.140 & 0.061 & 4.2 & $\pmb{\bm4e^{-5}}$ & 0.010 & $2e^{-4}$ & 0.053 & 0.470 & 0.530 & 0.177 & 4.0 \\
GRAN & 0.010 & \textbf{0.003} & 0.028 & 0.310 & \textbf{0.015} & 0.080 & 0.074 & 3.7 & 0.005 & 0.006 & $2e^{-4}$ & 0.023 & 0.172 & 0.660 & 0.144 & 3.7 \\
GNF & 0.637 & -- & 0.340 & -- & -- & -- & 0.489 & 7.3 & 0.046 & -- & 0.189 & -- & -- & -- & 0.117 & 7.5 \\
GraphAF & 0.262 & -- & 0.069 & 0.074 & -- & -- & 0.135 & 6.3 & $2.5e^{-3}$ & -- & 0.011 & 0.024 & -- & -- & 0.013 & 6.0 \\
GrAD-R & 0.009 & 0.510 & 0.420 & 0.420 & 0.150 & 0.500 & 0.335 & 6.2 & $3e^{-4}$ & 1.007 & 0.096 & 0.025 & 0.950 & 1.272 & 0.558 & 5.8 \\
GrAD-D & \textbf{0.005} & 0.310 & 0.001 & 0.050 & 0.027 & 0.053 & 0.074 & 2.8 & $1.3e^{-4}$ & 0.620 & $1e^{-5}$ & 0.025 & 0.075 & 1.170 & 0.315 &  3.7 \\
GrAD & \textbf{0.005} & 0.004 & 0.002 & \textbf{0.029} & 0.024 & \textbf{0.001} &  \textbf{0.011} & \textbf{1.7} & $1.2e^{-4}$ & \textbf{0.004} & \textbf{0} & 0.024 & \textbf{0.022} & \textbf{0.120} & \textbf{0.028} & \textbf{1.5} \\
\midrule
& \multicolumn{8}{c}{Spectra} & \multicolumn{8}{c}{Clustering Coefficient} \\
\midrule
GraphRNN & 0.110 & 0.040 & 0.053 & 0.015 & 0.100 & 0.014 & 0.055 & 4.2 & \textbf{0} & 0.960 & \textbf{0} & \textbf{0.045} & 0.650 & 0.130 & 0.298 & 2.3 \\
GraphRNN-S & 0.120 & 0.031 & 0.084 & 0.026 & 0.089 & 0.310 & 0.110 & 4.7 & \textbf{0} & 0.012 & 0.024 & 0.067 & 0.530 & \textbf{0.090} & 0.121 & 3.2 \\
GRAN & \textbf{0.032} & 0.013 & \textbf{0.032} & 0.023 & \textbf{0.040} & 0.026 & \textbf{0.028} & 2.2 & 0.005 & $\pmb{\bm4e^{-4}}$ & $2.3e^{-4}$ & 0.150 & 0.071 & 0.230 & 0.076 & 3.7 \\
GNF & 0.308 & -- & 0.256 & -- & -- & -- & 0.282 & 7.3 & 0.029 & -- & 0.134 & -- & -- & -- & 0.082 & 7.0 \\
GraphAF & 0.162 & -- & 0.229 & $7.6e^{-3}$ & -- & -- & 0.133 & 6.2 & 0.629 & -- & 0.274 & 0.055 & -- & -- & 0.319 & 6.5 \\
GrAD-R & 0.034 & 0.200 & 0.381 & 0.430 & 0.310 & 0.460 & 0.303 & 6.2 & $1.0e^{-3}$ & 1.010 & 0.230 & 0.058 & 0.770 & 0.820 & 0.482 & 5.5 \\
GrAD-D & \textbf{0.032} & 0.032 & 0.039 & 0.006 & 0.098 & 0.037 & 0.041 & 3.0 & $6e^{-5}$ & 1.430 & 0.002 & 0.058 & 0.086 & 0.235 & 0.302 & 4.2 \\
GrAD & \textbf{0.032} & \textbf{0.012} & 0.034 & $\pmb{\bm3.7e^{-3}}$ & 0.086 & $\pmb{\bm3.6e^{-3}}$ & 0.029 & \textbf{1.3} & \textbf{0} & 0.013 & 0.002 & 0.055 & \textbf{0.068} & 0.130 & \textbf{0.045} & \textbf{2.0} \\
\bottomrule
\end{tabular}
}
\caption{Comparison of GrAD against state-of-the-art models in terms of MMD on four different graph statistics. Lower is better. `--' indicates that training diverged or ran out of memory.}
\label{tab:quant}
\end{table*}

\section{Experiments}
\label{sec:exp}

We follow the general experimental setup of \citet{you2018graphrnn} and \citet{liao2019gran}. Details on datasets, metrics, baselines, and hyperparameters are provided in the supplement.

\textbf{Datasets.}
We use six datasets: four families of synthetic graphs (Cycles, Grid, Lobster, Community) and two sets of real graphs (Protein, Ego).

\textbf{Metrics.}
We quantitatively evaluate a graph generative model by comparing four different graph statistics~-- Degree, Clustering coefficients, Orbit, and Spectra~-- against the test set.

\textbf{Baselines.}
We compare GrAD to state-of-the-art sequential models~-- GraphRNN~\cite{you2018graphrnn} and GRAN~\cite{liao2019gran}~-- and normalizing flow models~-- GNF~\cite{liu2019graph} and GraphAF~\cite{shi2020graphaf}. We also report the performance of GraphRNN-S~\cite{you2018graphrnn} which generates all edges for each new node jointly. Further, we include the results of decoder-only versions of GrAD, dubbed GrAD-R and GrAD-D, in which the node embeddings are sampled from a Gaussian distribution. During training, the GrAD-R model does not learn latent codes and is fed with random codes while GrAD-D training is similar to GrAD.

\textbf{Results.}
The MMD evaluation on four graph statistics for all models trained on the six datasets is reported in Table~\ref{tab:quant}. In addition to MMD on each dataset and the mean across datasets, Table~\ref{tab:quant} also reports the average rank of each algorithm across datasets. For instance, if an algorithm achieves the 2nd best MMD on half of the datasets and the 3rd best MMD on the other half, its average rank is $2.5$. This is a robust summary measure of relative accuracy across multiple datasets.

GrAD achieves the best rank across datasets in each of the four measures. GrAD also has the lowest mean MMD by a factor of at least $1.5$ relative to the next best method in three of the four measures.
Samples from the trained GraphRNN, GRAN, and GrAD models are shown in Figure~\ref{fig:qual}. The GrAD samples more closely resemble the ground-truth graphs.

A trained GrAD can synthesize graphs of any size. Figure~\ref{fig:qual} illustrates this by showing graphs produced by GrAD for node cardinalities $N$ that are beyond the ranges observed in the training data: $200$ for Lobster, $200$ for Community, and $600$ for Protein. Even in this strong generalization regime, the synthesized graphs retain the qualitative characteristics of the training data. More qualitative and quantitative results are provided in the supplement.

\textbf{Runtime.}
Table~\ref{tab:inference} reports the runtime of our approach and the baselines. We use $K=1$ for our approach and GRAN. Runtime is measured on a GTX 1080 Ti and is averaged over 100 sampled graphs. GrAD is roughly 2x faster than GRAN on average, and more than an order of magnitude faster than GraphRNN.

\begin{table}[t]
\parbox{0.5\linewidth}{
\centering
\ra{1.05}
\resizebox{\linewidth}{!}{
\begin{tabular}{@{}l@{\hspace{1mm}}|c@{\hspace{2mm}}c@{\hspace{2mm}}c@{\hspace{2mm}}c@{\hspace{2mm}}c@{\hspace{2mm}}c|@{\hspace{2mm}}c@{}}
\toprule
Algorithm & Cycle & Grid & Lobs. & Comm. & Ego & Prot. & Mean \\
\midrule
GraphRNN & 0.63 & 8.60 & 2.20 & 6.10 & 39.0 & 5.30 & 10.3\\
GraphRNN-S & 0.08 & 0.07 & 0.28 & 0.12 & 0.31 & 0.50 & 0.23\\
GRAN & 0.45 & 2.25 & 0.44 & 0.85 & 2.17 & 3.48 & 1.60 \\
GrAD-D & 0.17 & 0.86 & 0.25 & 0.40 & 0.50 & 0.96 & 0.52\\
GrAD & 0.32  & 1.37 & 0.40& 0.62 & 0.81 & 1.70 & 0.87 \\
\bottomrule
\end{tabular}
}
\caption{Per-graph sampling time in seconds.}
\label{tab:inference}
}
\parbox{0.5\linewidth}{
\centering
\ra{1.05}
\resizebox{\linewidth}{!}{
\begin{tabular}{@{}l@{\hspace{1mm}}|c@{\hspace{2mm}}c@{\hspace{2mm}}c@{\hspace{2mm}}c|@{\hspace{2mm}}c@{\hspace{2mm}}c@{\hspace{2mm}}c@{\hspace{2mm}}c@{}}
\toprule
K & Degree & Orbit & Spectra & Clust. & Degree & Orbit & Spectra & Clust. \\
\midrule
 & \multicolumn{3}{c}{Stride=K} & & \multicolumn{3}{c}{Stride=1} \\
\midrule
2 & 0.007 & 0.001 & 0.064 & 0.13 & 0.001 & $2e^{-5}$ & 0.036 & 0.001\\
4 & 0.094 & 0.004 & 0.220 & 0.22 & 0.004 & $7e^{-4}$ & 0.051 & 0.002\\
8 & 0.110 & 0.016 & 0.270 & 0.48 & 0.006 & $4e^{-4}$ & 0.056 & 0.055\\
\bottomrule
\end{tabular}
}
\caption{Effect of block size $K$ on accuracy, measured on the Lobster dataset.}
\label{tab:efficiency}
}
\end{table}

\begin{figure}[!t]
\centering
\includegraphics[width=\linewidth]{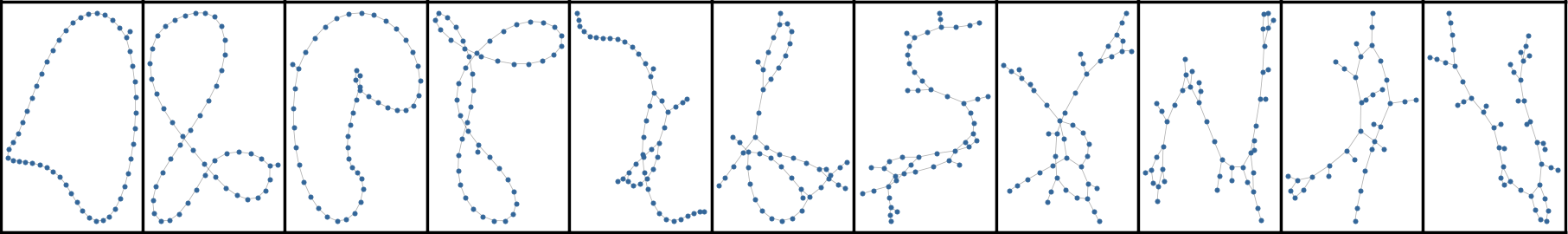}
\caption{A single GrAD model can be controlled to interpolate between the characteristics of different graph families. Most extreme graphs were sampled using one-hot code for cycles and lobster respectively. Intermediate graphs were synthesized using convex combination of these one-hot codes.}
\label{fig:interp}
\end{figure}

\textbf{Ablation studies.}
We conduct a number of ablation studies using the Lobster dataset. Table~\ref{tab:efficiency} quantifies the effect of block size $K$ on accuracy. As expected, increasing $K$ reduces accuracy. Applying strided sampling~\citep{liao2019gran} with stride 1 ameliorates the degradation. Table~\ref{tab:ablation} shows the accuracy of a number of ablated forms of our model. We begin by replacing GA, the basic processing block in our model, with attention-based GNN~\cite{liao2019gran}, GAT~\cite{velickovic2018graph}, or full self-attention~\citep{vaswani2017attention}.
\begin{wraptable}{r}{0.4\textwidth}
\centering
\resizebox{\linewidth}{!}{
\begin{tabular}{@{}l@{\hspace{1mm}}|c@{\hspace{2mm}}c@{\hspace{2mm}}c@{\hspace{2mm}}c@{}}
\toprule
Condition & Degree & Orbit & Spectra & Clust. \\
\midrule
 - & 0.002 & \textbf{0} & 0.034 & 0.002 \\
\midrule
Full Atten. ($\F_{\bm\Phi}$) & 0.002 & $1e^{-5}$ & 0.038 & 0.003 \\
Full Atten. ($\D_\btheta$) & 0.022 & 0.005 & 0.100 & 0.200 \\
GNN Atten. & 0.014 & $3e^{-4}$ & 0.048 & $\pmb{\bm3e^{-4}}$ \\
GAT & 0.006 & $2e^{-4}$ & 0.056 & 0.030 \\
\midrule
$C=1$ & 0.004 & $1e^{-4}$ & 0.061 & 0.007 \\
$C=40$ & \textbf{0.001} & $4e^{-5}$ & 0.033 & 0.006 \\
\midrule
$M=1$ & 0.008 & $5e^{-4}$ & 0.086 & 0.024 \\
$M=5$ & 0.002 & $3e^{-5}$ & 0.038 & $7e^{-4}$ \\
\midrule
$d=16$ & 0.003 & $1e^{-4}$ & 0.049 & 0.006 \\
$d=64$ & \textbf{0.001} & $2e^{-5}$ & \textbf{0.030} & 0.004 \\
\bottomrule
\end{tabular}
}
\caption{Ablation experiments on the Lobster dataset.}
\label{tab:ablation}
\end{wraptable}
We note that the GNN layers used in GRAN involve complex processing through GRU units while GA only uses MLPs. Our GA layer performs better than attentional GNN and GAT. Using full self-attention for node embeddings in the decoder substantially undermines accuracy. Using full self-attention in the density model marginally degrades overall performance.

Next we experiment with variants of the parameterized representation used for sampling edges.
If we replace the mixture representations with single Bernoulli distributions ($C=1$), performance declines. On the other hand, doubling the number of mixture components (from 20, our default, to 40) yields a marginal increase in accuracy. We observe similar trends in varying the number of message passing steps $M$ and the latent dimensionality $d$.

\textbf{Graph interpolation.}
GrAD can leverage the code sets used as input to the decoder to interpolate between the characteristics of different graph families. To demonstrate this, we train a single GrAD on two different datasets: Cycles and Lobster. During training, graph embeddings are concatenated with a one-hot code indicating which family the graph was drawn from. Figure~\ref{fig:interp} illustrates samples synthesized by this trained model when provided with different convex combinations of these one-hot codes.

\section{Conclusion}

We presented a new sequential generative model for graphs. At the core of our model is an auto-decoder: a generator that is trained by optimizing its parameters jointly with the latent codes that serve as its input. We rectified a deficiency in the auto-decoder framework by augmenting it with a flow-based density model that synthesizes latent codes that match the code distribution obtained during training. The resulting model is both more accurate and faster than the prior state of the art on large graphs. Our work contributes to the empirical modeling of network structures, which may assist a variety of science and engineering disciplines.

\small
\bibliographystyle{plainnat}
\bibliography{submission}

\begin{thebibliography}{49}
\providecommand{\natexlab}[1]{#1}
\providecommand{\url}[1]{\texttt{#1}}
\expandafter\ifx\csname urlstyle\endcsname\relax
  \providecommand{\doi}[1]{doi: #1}\else
  \providecommand{\doi}{doi: \begingroup \urlstyle{rm}\Url}\fi

\bibitem[Ba et~al.(2016)Ba, Kiros, and Hinton]{ba2016layer}
Jimmy~Lei Ba, Jamie~Ryan Kiros, and Geoffrey~E Hinton.
\newblock Layer normalization.
\newblock \emph{arXiv:1607.06450}, 2016.

\bibitem[Barab{\'a}si and Albert(1999)]{Barabasi1999}
Albert-L{\'a}szl{\'o} Barab{\'a}si and R{\'e}ka Albert.
\newblock Emergence of scaling in random networks.
\newblock \emph{Science}, 286, 1999.

\bibitem[Bello et~al.(2019)Bello, Zoph, Vaswani, Shlens, and
  Le]{bello2019attention}
Irwan Bello, Barret Zoph, Ashish Vaswani, Jonathon Shlens, and Quoc~V Le.
\newblock Attention augmented convolutional networks.
\newblock In \emph{ICCV}, 2019.

\bibitem[Bishop(2006)]{bishop2006pattern}
Christopher~M Bishop.
\newblock \emph{Pattern Recognition and Machine Learning}.
\newblock Springer, 2006.

\bibitem[Bojanowski et~al.(2018)Bojanowski, Joulin, Lopez-Pas, and
  Szlam]{bojanowski2018optimizing}
Piotr Bojanowski, Armand Joulin, David Lopez-Pas, and Arthur Szlam.
\newblock Optimizing the latent space of generative networks.
\newblock In \emph{ICML}, 2018.

\bibitem[Brock et~al.(2019)Brock, Donahue, and Simonyan]{brock2019large}
Andrew Brock, Jeff Donahue, and Karen Simonyan.
\newblock Large scale {GAN} training for high fidelity natural image synthesis.
\newblock In \emph{ICLR}, 2019.

\bibitem[Broido and Clauset(2019)]{BroidoClauset2019}
Anna~D. Broido and Aaron Clauset.
\newblock Scale-free networks are rare.
\newblock \emph{Nature Communications}, 10, 2019.

\bibitem[Dai et~al.(2013)Dai, Ding, and Wahba]{dai2013multivariate}
Bin Dai, Shilin Ding, and Grace Wahba.
\newblock Multivariate {Bernoulli} distribution.
\newblock \emph{Bernoulli}, 19\penalty0 (4):\penalty0 1465--1483, 2013.

\bibitem[Dai et~al.(2019)Dai, Yang, Yang, Carbonell, Le, and
  Salakhutdinov]{dai2019transformer}
Zihang Dai, Zhilin Yang, Yiming Yang, Jaime Carbonell, Quoc~V. Le, and Ruslan
  Salakhutdinov.
\newblock {Transformer-XL}: Attentive language models beyond a fixed-length
  context.
\newblock In \emph{ACL}, 2019.

\bibitem[De~Cao and Kipf(2018)]{de2018molgan}
Nicola De~Cao and Thomas Kipf.
\newblock {MolGAN}: An implicit generative model for small molecular graphs.
\newblock In \emph{ICML Workshops}, 2018.

\bibitem[Devlin et~al.(2019)Devlin, Chang, Lee, and Toutanova]{Devlin2018}
Jacob Devlin, Ming-Wei Chang, Kenton Lee, and Kristina Toutanova.
\newblock {BERT}: Pre-training of deep bidirectional transformers for language
  understanding.
\newblock In \emph{NAACL-HLT}, 2019.

\bibitem[Dinh et~al.(2014)Dinh, Krueger, and Bengio]{dinh2014nice}
Laurent Dinh, David Krueger, and Yoshua Bengio.
\newblock {NICE}: Non-linear independent components estimation.
\newblock \emph{arXiv:1410.8516}, 2014.

\bibitem[Dinh et~al.(2017)Dinh, Sohl-Dickstein, and Bengio]{45819}
Laurent Dinh, Jascha Sohl-Dickstein, and Samy Bengio.
\newblock Density estimation using {Real} {NVP}.
\newblock In \emph{ICLR}, 2017.

\bibitem[Dobson and Doig(2003)]{dobson2003distinguishing}
Paul~D Dobson and Andrew~J Doig.
\newblock Distinguishing enzyme structures from non-enzymes without alignments.
\newblock \emph{Journal of Molecular Biology}, 330\penalty0 (4), 2003.

\bibitem[Erd{\H{o}}s and R{\'e}nyi(1959)]{erdds1959random}
Paul Erd{\H{o}}s and Alfr{\'e}d R{\'e}nyi.
\newblock On random graphs {I}.
\newblock \emph{Publ. Math. Debrecen}, 6, 1959.

\bibitem[Fan and Cheng(2018)]{fan2018matrix}
Jicong Fan and Jieyu Cheng.
\newblock Matrix completion by deep matrix factorization.
\newblock \emph{Neural Networks}, 98, 2018.

\bibitem[Germain et~al.(2015)Germain, Gregor, Murray, and
  Larochelle]{germain2015made}
Mathieu Germain, Karol Gregor, Iain Murray, and Hugo Larochelle.
\newblock {MADE}: Masked autoencoder for distribution estimation.
\newblock In \emph{ICML}, 2015.

\bibitem[Jackson and Rogers(2007)]{JacksonRogers2007}
Matthew~O. Jackson and Brian~W. Rogers.
\newblock Meeting strangers and friends of friends: How random are social
  networks?
\newblock \emph{The American Economic Review}, 97, 2007.

\bibitem[Jin et~al.(2018)Jin, Barzilay, and Jaakkola]{jin2018junction}
Wengong Jin, Regina Barzilay, and Tommi Jaakkola.
\newblock Junction tree variational autoencoder for molecular graph generation.
\newblock In \emph{ICML}, 2018.

\bibitem[Kingma and Dhariwal(2018)]{kingma2018glow}
Durk~P Kingma and Prafulla Dhariwal.
\newblock Glow: Generative flow with invertible 1x1 convolutions.
\newblock In \emph{Advances in Neural Information Processing Systems}, 2018.

\bibitem[Kingma et~al.(2016)Kingma, Salimans, Jozefowicz, Chen, Sutskever, and
  Welling]{kingma2016improved}
Durk~P Kingma, Tim Salimans, Rafal Jozefowicz, Xi~Chen, Ilya Sutskever, and Max
  Welling.
\newblock Improved variational inference with inverse autoregressive flow.
\newblock In \emph{Advances in Neural Information Processing Systems}, 2016.

\bibitem[Kipf and Welling(2016)]{kipf2016variational}
Thomas~N Kipf and Max Welling.
\newblock Variational graph auto-encoders.
\newblock \emph{arXiv:1611.07308}, 2016.

\bibitem[Leskovec et~al.(2010)Leskovec, Chakrabarti, Kleinberg, Faloutsos, and
  Ghahramani]{leskovec2010kronecker}
Jure Leskovec, Deepayan Chakrabarti, Jon Kleinberg, Christos Faloutsos, and
  Zoubin Ghahramani.
\newblock Kronecker graphs: An approach to modeling networks.
\newblock \emph{Journal of Machine Learning Research}, 11, 2010.

\bibitem[Li et~al.(2018)Li, Vinyals, Dyer, Pascanu, and
  Battaglia]{li2018learning}
Yujia Li, Oriol Vinyals, Chris Dyer, Razvan Pascanu, and Peter Battaglia.
\newblock Learning deep generative models of graphs.
\newblock \emph{arXiv:1803.03324}, 2018.

\bibitem[Liao et~al.(2019)Liao, Li, Song, Wang, Nash, Hamilton, Duvenaud,
  Urtasun, and Zemel]{liao2019gran}
Renjie Liao, Yujia Li, Yang Song, Shenlong Wang, Charlie Nash, William~L.
  Hamilton, David Duvenaud, Raquel Urtasun, and Richard Zemel.
\newblock Efficient graph generation with graph recurrent attention networks.
\newblock In \emph{Advances in Neural Information Processing Systems}, 2019.

\bibitem[Lima-Mendez and van Helden(2009)]{Lima-Mendez2009}
Gipsi Lima-Mendez and Jacques van Helden.
\newblock The powerful law of the power law and other myths in network biology.
\newblock \emph{Molecular BioSystems}, 5, 2009.

\bibitem[Liu et~al.(2019)Liu, Kumar, Ba, Kiros, and Swersky]{liu2019graph}
Jenny Liu, Aviral Kumar, Jimmy Ba, Jamie Kiros, and Kevin Swersky.
\newblock Graph normalizing flows.
\newblock In \emph{Advances in Neural Information Processing Systems}, 2019.

\bibitem[Liu et~al.(2018)Liu, Allamanis, Brockschmidt, and
  Gaunt]{liu2018constrained}
Qi~Liu, Miltiadis Allamanis, Marc Brockschmidt, and Alexander Gaunt.
\newblock Constrained graph variational autoencoders for molecule design.
\newblock In \emph{Advances in Neural Information Processing Systems}, 2018.

\bibitem[Ma et~al.(2018)Ma, Chen, and Xiao]{ma2018constrained}
Tengfei Ma, Jie Chen, and Cao Xiao.
\newblock Constrained generation of semantically valid graphs via regularizing
  variational autoencoders.
\newblock In \emph{Advances in Neural Information Processing Systems}, 2018.

\bibitem[Madhawa et~al.(2019)Madhawa, Ishiguro, Nakago, and
  Abe]{madhawa2019graphnvp}
Kaushalya Madhawa, Katushiko Ishiguro, Kosuke Nakago, and Motoki Abe.
\newblock {GraphNVP}: An invertible flow model for generating molecular graphs.
\newblock \emph{arXiv:1905.11600}, 2019.

\bibitem[Mandt et~al.(2017)Mandt, Hoffman, and Blei]{mandt2017stochastic}
Stephan Mandt, Matthew~D Hoffman, and David~M Blei.
\newblock Stochastic gradient descent as approximate {Bayesian} inference.
\newblock \emph{Journal of Machine Learning Research}, 18:\penalty0 4873--4907,
  2017.

\bibitem[Newman(2018)]{Newman2018}
Mark Newman.
\newblock \emph{Networks}.
\newblock Oxford University Press, 2nd edition, 2018.

\bibitem[Papamakarios et~al.(2017)Papamakarios, Pavlakou, and
  Murray]{papamakarios2017masked}
George Papamakarios, Theo Pavlakou, and Iain Murray.
\newblock Masked autoregressive flow for density estimation.
\newblock In \emph{Advances in Neural Information Processing Systems}, 2017.

\bibitem[Papamakarios et~al.(2019)Papamakarios, Nalisnick, Rezende, Mohamed,
  and Lakshminarayanan]{papamakarios2019normalizing}
George Papamakarios, Eric Nalisnick, Danilo~Jimenez Rezende, Shakir Mohamed,
  and Balaji Lakshminarayanan.
\newblock Normalizing flows for probabilistic modeling and inference.
\newblock \emph{arXiv:1912.02762}, 2019.

\bibitem[Park et~al.(2019)Park, Florence, Straub, Newcombe, and
  Lovegrove]{park2019deepsdf}
Jeong~Joon Park, Peter Florence, Julian Straub, Richard Newcombe, and Steven
  Lovegrove.
\newblock {DeepSDF}: Learning continuous signed distance functions for shape
  representation.
\newblock In \emph{CVPR}, 2019.

\bibitem[Prenger et~al.(2019)Prenger, Valle, and Catanzaro]{Prenger2019}
Ryan Prenger, Rafael Valle, and Bryan Catanzaro.
\newblock {WaveGlow}: A flow-based generative network for speech synthesis.
\newblock In \emph{ICASSP}, 2019.

\bibitem[Ramachandran et~al.(2019)Ramachandran, Parmar, Vaswani, Bello,
  Levskaya, and Shlens]{ramachandran2019stand}
Prajit Ramachandran, Niki Parmar, Ashish Vaswani, Irwan Bello, Anselm Levskaya,
  and Jonathon Shlens.
\newblock Stand-alone self-attention in vision models.
\newblock In \emph{Advances in Neural Information Processing Systems}, 2019.

\bibitem[Sen et~al.(2008)Sen, Namata, Bilgic, Getoor, Galligher, and
  Eliassi-Rad]{sen2008collective}
Prithviraj Sen, Galileo Namata, Mustafa Bilgic, Lise Getoor, Brian Galligher,
  and Tina Eliassi-Rad.
\newblock Collective classification in network data.
\newblock \emph{AI Magazine}, 29, 2008.

\bibitem[Shi et~al.(2020)Shi, Xu, Zhu, Zhang, Zhang, and Tang]{shi2020graphaf}
Chence Shi, Minkai Xu, Zhaocheng Zhu, Weinan Zhang, Ming Zhang, and Jian Tang.
\newblock {GraphAF}: A flow-based autoregressive model for molecular graph
  generation.
\newblock In \emph{ICLR}, 2020.

\bibitem[Simonovsky and Komodakis(2018)]{Simonovsky2018}
Martin Simonovsky and Nikos Komodakis.
\newblock {GraphVAE}: Towards generation of small graphs using variational
  autoencoders.
\newblock In \emph{ICANN}, 2018.

\bibitem[Tan and Mayrovouniotis(1995)]{tan1995reducing}
Shufeng Tan and Michael~L Mayrovouniotis.
\newblock Reducing data dimensionality through optimizing neural network
  inputs.
\newblock \emph{AIChE Journal}, 41, 1995.

\bibitem[Vaswani et~al.(2017)Vaswani, Shazeer, Parmar, Uszkoreit, Jones, Gomez,
  Kaiser, and Polosukhin]{vaswani2017attention}
Ashish Vaswani, Noam Shazeer, Niki Parmar, Jakob Uszkoreit, Llion Jones,
  Aidan~N Gomez, {\L}ukasz Kaiser, and Illia Polosukhin.
\newblock Attention is all you need.
\newblock In \emph{Advances in Neural Information Processing Systems}, 2017.

\bibitem[Veli{\v{c}}kovi{\'{c}} et~al.(2018)Veli{\v{c}}kovi{\'{c}}, Cucurull,
  Casanova, Romero, Li{\`{o}}, and Bengio]{velickovic2018graph}
Petar Veli{\v{c}}kovi{\'{c}}, Guillem Cucurull, Arantxa Casanova, Adriana
  Romero, Pietro Li{\`{o}}, and Yoshua Bengio.
\newblock Graph attention networks.
\newblock In \emph{ICLR}, 2018.

\bibitem[Wasserman and Pattison(1996)]{wasserman1996logit}
Stanley Wasserman and Philippa Pattison.
\newblock Logit models and logistic regressions for social networks: I. {An}
  introduction to {Markov} graphs and p*.
\newblock \emph{Psychometrika}, 61\penalty0 (3), 1996.

\bibitem[Watts and Strogatz(1998)]{WattsStrogatz1998}
Duncan~J. Watts and Steven~H. Strogatz.
\newblock Collective dynamics of `small-world' networks.
\newblock \emph{Nature}, 393, 1998.

\bibitem[Yang et~al.(2019)Yang, Dai, Yang, Carbonell, Salakhutdinov, and
  Le]{Yang2019xlnet}
Zhilin Yang, Zihang Dai, Yiming Yang, Jaime Carbonell, Ruslan Salakhutdinov,
  and Quoc~V. Le.
\newblock {XLNet}: Generalized autoregressive pretraining for language
  understanding.
\newblock In \emph{Advances in Neural Information Processing Systems}, 2019.

\bibitem[You et~al.(2018{\natexlab{a}})You, Liu, Ying, Pande, and
  Leskovec]{you2018graph}
Jiaxuan You, Bowen Liu, Zhitao Ying, Vijay Pande, and Jure Leskovec.
\newblock Graph convolutional policy network for goal-directed molecular graph
  generation.
\newblock In \emph{Advances in Neural Information Processing Systems},
  2018{\natexlab{a}}.

\bibitem[You et~al.(2018{\natexlab{b}})You, Ying, Ren, Hamilton, and
  Leskovec]{you2018graphrnn}
Jiaxuan You, Rex Ying, Xiang Ren, William Hamilton, and Jure Leskovec.
\newblock {GraphRNN}: Generating realistic graphs with deep auto-regressive
  models.
\newblock In \emph{ICML}, 2018{\natexlab{b}}.

\bibitem[Zhao et~al.(2020)Zhao, Jia, and Koltun]{Zhao2020}
Hengshuang Zhao, Jiaya Jia, and Vladlen Koltun.
\newblock Exploring self-attention for image recognition.
\newblock In \emph{CVPR}, 2020.

\end{thebibliography}

\appendix
\section*{Appendices}
\addcontentsline{toc}{section}{Appendices}
\renewcommand{\thesubsection}{\Alph{subsection}}

\subsection{Datasets}
We use six datasets to conduct a thorough comparison to the state of the art: four families of synthetic graphs and two sets of real-world graphs. (1) Cycles: 95 standard circular graphs with $5 \leq \abs{V} \leq 100$. (2) Grid: 121 standard 2D grid graphs with $100 \leq \abs{V} \leq 400$. (3) Lobster: 100 graphs with $10 \leq \abs{V} \leq 100$. A lobster graph is a tree wherein all the vertices are within distance two of a central path. (4) Community: 510 two-community graphs with $60 \leq \abs{V} \leq 160$. These are generated using the Erd{\H{o}}s-R{\'e}nyi model~\cite{erdds1959random} with each community containing half the nodes and intra-community edge probability set to $p=0.3$. Following this, $0.05\abs{V}$ inter-community edges are added uniformly. (5) Protein: 918 protein graphs~\cite{dobson2003distinguishing} with $100 \leq \abs{V} \leq 500$. Each graph represents a distinct protein molecule where nodes imply amino acids and two nodes are connected if the corresponding amino acids are separated by less than 6 Angstrom. (6) Ego: 757 graphs with $50 \leq \abs{V} \leq 399$. These are extracted from the Citeseer network~\cite{sen2008collective} constrained to 3-hop connectivity, with each node representing a document and edges corresponding to citation relationships. Following the protocol in~\citet{you2018graphrnn}, we generate a random split with $80\%$ of each dataset used for training and the rest used for testing.

\subsection{Metrics}
We quantitatively evaluate a graph generative model by comparing the distribution of various graph statistics in a set of graphs sampled from the model and the corresponding test set. We follow the evaluation procedure recommended by~\citet{liao2019gran} and measure four different graph statistics. (1) Degree: degree counts of all nodes in a graph. (2) Clustering coefficients. (3) Orbit: number of occurrence of all orbit in a graphlet with 4 nodes. (4) Spectra: distribution of quantized eigenvalues of normalized graph Laplacian. This is followed by computing the squared maximum mean discrepancy (MMD) distance for each of these statistics eith respect to the corresponding statistics in the test set. MMD is a measure of statistical dissimilarity of two sets of samples. Lower MMD suggests better fit to the data. The Spectra metric captures global graph properties while the other three metrics focus on local properties.

\subsection{Baselines}
We compare the performance of GrAD to the state-of-the-art sequential models, GraphRNN~\cite{you2018graphrnn} and GRAN~\cite{liao2019gran}. We also report the performance of GraphRNN-S~\cite{you2018graphrnn} which generates all edges for each new node jointly: this model is faster but less accurate than GraphRNN. Further, we draw comparison to state-of-the-art normalizing flow models: GNF~\cite{liu2019graph} and GraphAF~\cite{shi2020graphaf}. All models were retrained using the implementations provided by the authors. \citet{you2018graphrnn} and ~\citet{liu2019graph} use a fixed setting across all datasets and we follow their example. \citet{liao2019gran} report results with different settings for each dataset. For instance, they use different hidden dimensionality, batch size, and total number of training epochs for different datasets. For datasets used by \citet{liao2019gran}, we use their suggested best settings for their model. For other datasets, we report results on the model with the highest capacity. GraphAF uses node as well as edge types as feature input and report results only on small molecular graphs. We scale to standard datasets using their suggested settings for molecular graph generation. We consider rows of adjacency matrix as node features while for edge feature, two edge types are created: connected edges and non-connected edges. The maximum BFS depth is set to be dataset-specific.

We do not report results for GraphVAE, and DeepGMG, as they do not scale to the large datasets used to evaluate GraphRNN, GRAN, and our work.

\subsection{Experimental details}
For our model (GrAD), we use fixed settings across all datasets. We fix the node embedding dimensionality to $d=32$. The GA layer employs $8$ heads with hidden dimensionality $d_S$ set to $16$ for the decoder and $10$ for the density model. The number of message passing steps for decoder and flow is $M=2$ and $R=9$ respectively. The latent codes are updated with SGD optimizer with constant learning rate of $\delta=0.1$. The decoder and flow models are trained using Adam optimizer. For decoder, the initial learning rate is set to $\tau=5e^{-5}$ which is decreased by a factor of $0.3$ at every one-third of total epochs. Whereas the flow model is trained with an exponentially decaying learning rate starting with $1e^{-3}$. For every decoder updates the latent codes are updated twice. We train decoder and flow model for total of 500 and 800 epochs respectively. Due to limited training set, we avoid overfitting in flow model by adding random gaussian noise to input latent codes. Similar strategy is employed for training decoder model on small datasets such as Lobster and Cycles. At inference time, the flow model samples latent codes from Gaussian distribution using conservative $\sigma=0.7$.

Unless specified, all results are reported with block size $K=1$, $C=20$ mixtures, and breadth-first search (BFS) node ordering. We also report results on a decoder-only versions of GrAD, dubbed GrAD-R and GrAD-D, in which the node embeddings are sampled from a Gaussian distribution. (This is an ablation condition that evaluates the contribution of the density model $\F_{\bm\Phi}$). During training, the GrAD-R model do not learn latents and it is fed with random codes while GrAD-D jointly trains for latents as in GrAD. Moreover, to make a fair comparison, we train GrAD-R models for $3\times$ epochs and hence maintaining the similar training budget.

\subsection{Evaluation of out-of-distribution samples}
\begin{figure}[!htb]
\centering
\subfloat[Lobster: 150]{\includegraphics[width=0.20\textwidth]{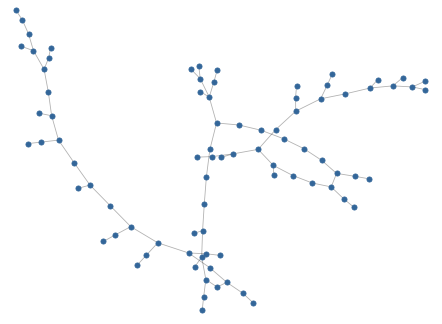}}
\subfloat[Lobster: 200]{\includegraphics[width=0.20\textwidth]{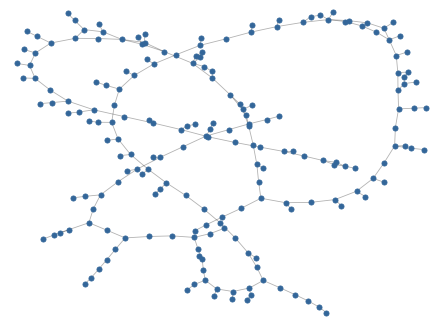}}
\subfloat[Lobster: 400]{\includegraphics[width=0.20\textwidth]{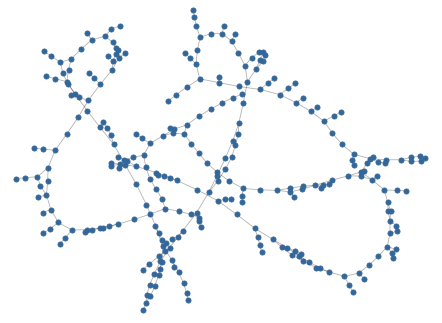}}
\subfloat[Lobster: 600]{\includegraphics[width=0.20\textwidth]{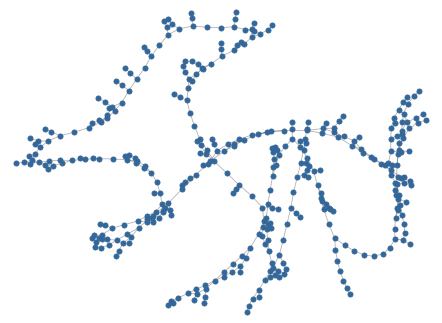}}
\subfloat[Cycles]{\includegraphics[width=0.20\textwidth]{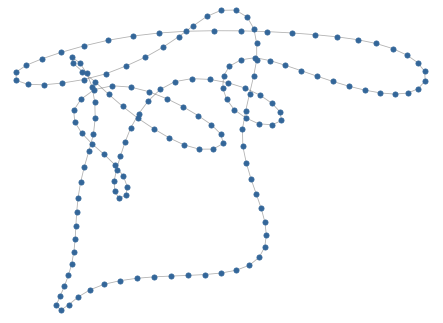}}
\\
\subfloat[Comm: 200]{\includegraphics[width=0.20\textwidth]{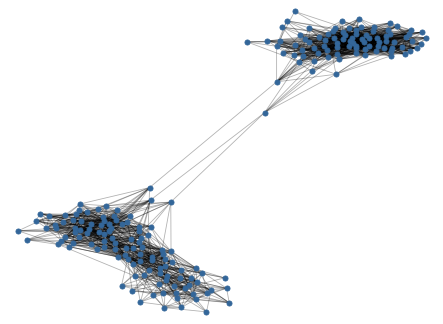}}
\subfloat[Comm: 300]{\includegraphics[width=0.20\textwidth]{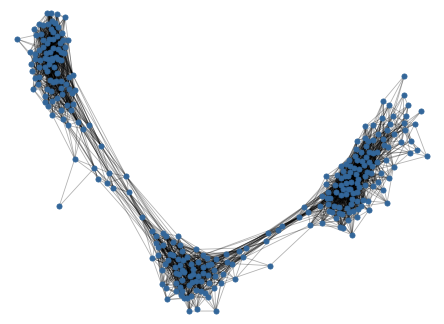}}
\subfloat[Comm: 400]{\includegraphics[width=0.20\textwidth]{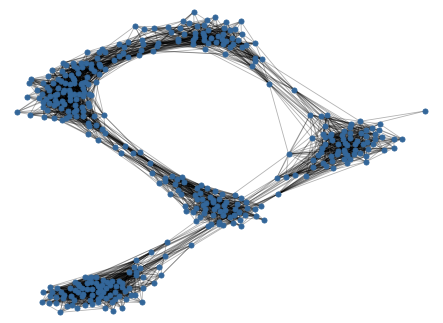}}
\subfloat[Comm: 500]{\includegraphics[width=0.20\textwidth]{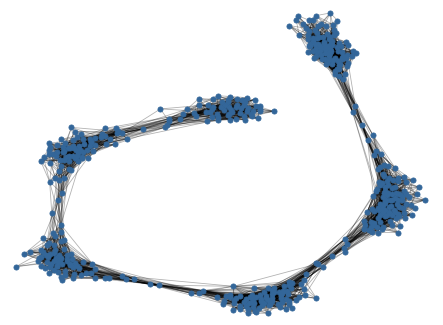}}
\subfloat[Grid]{\includegraphics[width=0.20\textwidth]{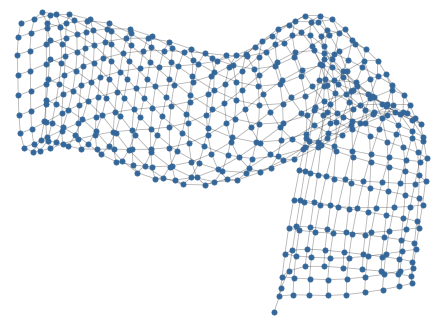}}
\caption{Out-of-distribution graphs sampled at various input graph size.}
\label{fig:ood}
\end{figure}

As discussed in section~\ref{sec:exp}, GrAD can straightaway generate out-of-distribution (OOD) samples. Unlike the prior state-of-the-art models such as GraphRNN and GRAN, this possibility arises in GrAD because the dimension of latent codes is not binded to the maximum training graph size. Few OOD graphs were displayed in last row of Figure~\ref{fig:qual}. Given that the synthetic graphs data (cycles, grid, lobster, community) can be easily synthesized for any number of given nodes, we try evaluating the fidelity of OOD samples against these set of OOD ground truth graphs using quantitative metrics. In comparison to in-distribution sampling results in Table~\ref{tab:quant}, the results for OOD sampling summarized in Table~\ref{tab:ood} suggests that GrAD model generalize very well to OOD samples. For random graphs, the evaluation is done at multiple fixed number of nodes and one can observe the changing dynamics in Figure~\ref{fig:ood}. For community data, the model tend to generates more communities with increasing graph size. However, model keeps the intrinsic structure of underlying dataset intact.
\begin{table}[!h]
\centering
\ra{1.05}
\resizebox{0.75\linewidth}{!}{
\begin{tabular}{@{}c@{\hspace{2mm}}|c@{\hspace{2mm}}c@{\hspace{2mm}}c@{\hspace{2mm}}c|@{\hspace{2mm}}c|@{\hspace{2mm}}c@{\hspace{2mm}}c@{\hspace{2mm}}c@{\hspace{2mm}}c@{}}
\toprule
Nodes & Degree & Orbit & Spectra & Clust. & Nodes & Degree & Orbit & Spectra & Clust. \\
\midrule
 & \multicolumn{3}{c}{Lobster} & & & \multicolumn{3}{c}{Community} \\
\midrule
150 & 0.008 & $1e^{-4}$ & 0.042 & 0.006 & 200 & 0.400 & 0.019 & 0.030 & 0.340\\
200 & 0.008 & $3e^{-4}$ & 0.034 & 0.008 & 300 & 0.690 & 0.021 & 0.069 & 0.570\\
400 & 0.012 & $18e^{-4}$ & 0.035 & 0.025 & 400 & 0.740 & 0.020 & 0.107 & 0.750\\
600 & 0.014 & $48e^{-4}$ & 0.042 & 0.056 & 500 & 0.760 & 0.021 & 0.142 & 0.830\\
\midrule
 & \multicolumn{3}{c}{Cycles} & & & \multicolumn{3}{c}{Grid} \\
\midrule
150-200 & $7e^{-4}$ & 0 & 0.047 & $1e^{-4}$ & 400-625 & 0.009 & 0.010 & 0.012 & 0.002\\
\bottomrule
\end{tabular}
}
\caption{Quantitative evaluation of out-of-distribution samples.}
\label{tab:ood}
\end{table}

\subsection{Ablation: Canonical Ordering}
Like GraphRNN, we use a single canonical ordering, BFS, for all our experiments. GRAN, which proposes framework for variable canonical ordering, also report results using single DFS ordering in their implementation. Determining the optimal ordering is NP-hard, and learning a good ordering is also difficult. GrAD model works best with BFS node ordering. This is because, unlike DFS and other orderings, BFS ordering bounds the dependency distance between nodes. This inturn alleviates shortcoming of sequential training process. Nevertheless, the performance is graph specific. Table~\ref{tab:ord} summarise the results achieved on lobster and community graphs when trained with different canonical ordering. On lobster, different canonical ordering works poorly than BFS while for community graphs the performance is equivalent to BFS.
\begin{table}[!h]
\centering
\ra{1.05}
\resizebox{0.65\linewidth}{!}{
\begin{tabular}{@{}c@{\hspace{2mm}}|c@{\hspace{2mm}}c@{\hspace{2mm}}c@{\hspace{2mm}}c|@{\hspace{2mm}}c@{\hspace{2mm}}c@{\hspace{2mm}}c@{\hspace{2mm}}c@{}}
\toprule
Nodes & Degree & Orbit & Spectra & Clust. & Degree & Orbit & Spectra & Clust. \\
\midrule
 & \multicolumn{3}{c}{Lobster} & & \multicolumn{3}{c}{Community} \\
\midrule
BFS & 0.002 & 0 & 0.034 & 0.002 & 0.029 & 0.024 & 0.004 & 0.055\\
\midrule
DFS & 0.100 & 0.680 & 0.340 & 0.680 & 0.019 & 0.025 & 0.004 & 0.054\\
Default & 0.140 & 0.029 & 0.330 & 0.005 & 0.037 & 0.025 & 0.010 & 0.059\\
Degree & 0.200 & 0.028 & 0.310 & 0.004 & 0.117 & 0.031 & 0.045 & 0.086\\
K-core & 0.130 & 0.022 & 0.270 & 0.019 & 0.029 & 0.025 & 0.014 & 0.059\\
\bottomrule
\end{tabular}
}
\caption{Effect of different canonical ordering on accuracy, measured on lobster and community datasets.}
\label{tab:ord}
\end{table}

\subsection{Training NLL}
Table~\ref{tab:nll} compares average negative log-likelihoods (NLL), $-\log p(\LL^\pi)$, on the training sets for two of our decoder-only models, GrAD-R and GrAD-D. We found that learning latents along with the model is useful in escaping local minima and thus leading to faster convergence with low NLL. This also leads to better accuracy measure for GrAD-D over GrAD-R (see Table~\ref{tab:quant}).
\begin{table}[!h]
\centering
\ra{1.05}
\resizebox{0.5\linewidth}{!}{
\begin{tabular}{@{}c@{\hspace{2mm}}|c@{\hspace{2mm}}c@{\hspace{2mm}}c@{\hspace{2mm}}c@{\hspace{2mm}}c@{\hspace{2mm}}c@{}}
\toprule
Model & Cycles & Grid & Lobs. & Comm. & Ego & Prot. \\
\midrule
GrAD-D & 0.286 & 0.520 & 3.271 & 800.0 & 323.7 & 30.0 \\
GrAD-R & 5.921 & 770.0 & 107.7 & 2100 & 1096 & 2140 \\
\bottomrule
\end{tabular}
}
\caption{Comparison of negative log-likelihood (NLL) for model trained with/without latents}
\label{tab:nll}
\end{table}

\subsection{Lobster Accuracy}
For sanity check, GRAN~\cite{liao2019gran} paper proposed a necessary condition to verify whether the sampled graph truly belongs to a family of Lobster or not. The input graph is a Lobster if it reduces to path graph when two-level of leaf nodes are discarded. If evaluated using this condition, the Lobster graph generation accuracy for various model is, GraphRNN: $95\%$, GraphRNN-S: $35\%$, GRAN: $66\%$ and GrAD: $70\%$. However, note that this is a coarse measure as it only satisfy necessary condition and not sufficiency. For instance, we find that the test passes for any simple path graph whereas an almost perfect Lobster with a node at distance 3 from central path fails. Latter is the case with GrAD.

\subsection{Comparison of training time and memory}
Due to the sequential training process, the per epoch training time of GrAD decoder model is much larger than GRAN (which trains subgraphs in parallel). However, the decoder training in GrAD converges in very few epochs than that utilized by GRAN. For example, the GRAN implementation uses 100K epochs to train on lobster dataset while GrAD trains within 500 epochs (fixed across all datasets). To make a fair comparison numerically, with batch size of 20 and 1080 Ti, GrAD trains on lobster within 8 hours consuming maximum of 6GB of GPU RAM while GRAN takes 45 hours with 10GB RAM.

\end{document}